\documentclass[twoside]{article}
\usepackage{etoolbox}
\usepackage[letterpaper, left=1in, right=1in, top=1in, bottom=1in]{geometry}
\usepackage[usenames,dvipsnames,svgnames]{xcolor}
\usepackage{parskip}

\usepackage{amsmath,amssymb,amsthm}
\usepackage{xspace}

\usepackage{url}
\usepackage{bm}
\usepackage{mathtools}

\usepackage{nicefrac}

\usepackage[round,sort&compress]{natbib}

\newcommand{\afterhead}{.}		%
\newcommand{\para}[1]{\paragraph{\S\;\textbf{#1\afterhead}}}

\usepackage[inline,shortlabels]{enumitem}		%
\setenumerate{itemsep=1pt,parsep=2pt,topsep=1pt,left=\parindent}
\setitemize{itemsep=1pt,parsep=2pt,topsep=1pt,left=\parindent}

\usepackage{hyperref}
\hypersetup{
final,
colorlinks=true,
linktocpage=true,
pdfstartview=FitH,
breaklinks=true,
pdfpagemode=UseNone,
pageanchor=true,
pdfpagemode=UseOutlines,
plainpages=false,
bookmarksnumbered,
bookmarksopen=false,
bookmarksopenlevel=1,
hypertexnames=false,
pdfhighlight=/O,
urlcolor=LinkColor,linkcolor=LinkColor,citecolor=LinkColor,	%
pdftitle={},
pdfauthor={},
pdfsubject={},
pdfkeywords={},
pdfcreator={pdfLaTeX},
pdfproducer={LaTeX with hyperref}
}

\usepackage[sort&compress,capitalize,nameinlink]{cleveref}		%
\crefname{assumption}{Assumption}{Assumptions}

\usepackage{thmtools}		%
\usepackage{thm-restate}		%
\usepackage{mdframed}
\declaretheorem[name=Theorem,numberwithin=section]{thm}

\usepackage[%
cal=cm,
]
{mathalfa}

\usepackage{quoting}			%
\quotingsetup{vskip=0pt}

\usepackage{acronym}		%
\newcommand{\acli}[1]{\textit{\acl{#1}}}		%
\newcommand{\acdef}[1]{\textit{\acl{#1}} \textup{(\acs{#1})}\acused{#1}}		%
\newcommand{\acdefp}[1]{\textit{\aclp{#1}} \textup{(\acsp{#1})}\acused{#1}}	%

\usepackage[labelfont={bf,small},labelsep=colon,font=small]{caption}	%

\usepackage[dvipsnames,svgnames]{xcolor}		%
\colorlet{MyRed}{FireBrick!50!Crimson}
\colorlet{MyBlue}{DodgerBlue!75!black}
\colorlet{MyGreen}{DarkGreen!85!black}
\colorlet{MyViolet}{DarkMagenta}

\colorlet{MyLightBlue}{DodgerBlue!20}
\colorlet{MyLightGreen}{MyGreen!20}

\colorlet{PrimalColor}{MyBlue}
\colorlet{PrimalFill}{MyLightBlue}
\colorlet{DualColor}{MyRed}

\colorlet{RevColor}{BlueViolet}
\colorlet{LinkColor}{MediumBlue}

\theoremstyle{plain}
\newtheorem{theorem}{Theorem}		%
\newtheorem*{theorem*}{Theorem}		%
\newtheorem{lemma}{Lemma}		%
\newtheorem{proposition}{Proposition}		%

\newtheorem{corollary}{Corollary}		%
\newtheorem*{corollary*}{Corollary}		%

\newtheorem{assumption}{Assumption}		%

\theoremstyle{definition}
\newtheorem{definition}{Definition}		%

\newtheorem*{definition*}{Definition}		%
\newtheorem*{assumption*}{Assumption}		%
\newtheorem*{blanket*}{Blanket assumption}		%
\newtheorem*{example*}{Example}		%

\theoremstyle{remark}
\newtheorem*{remark*}{Remark}		%

\DeclarePairedDelimiter{\nrm}{|}{|}
\DeclarePairedDelimiter{\Nrm}{\|}{\|}

\DeclarePairedDelimiter{\prn}{(}{)}
\DeclarePairedDelimiter{\crl}{\{}{\}}
\DeclarePairedDelimiter{\brk}{[}{]}
\DeclarePairedDelimiter{\ang}{\langle}{\rangle}

\makeatletter
\newcommand\inner{%
  \@ifstar{\innerstar}{\innernostar}}

\newcommand{\innernostar}[1]{\ang{#1}}
\newcommand{\innerstar}[1]{\ang*{#1}}
\makeatother

\newcommand{\Q}{\mathbb{Q}}		%
\newcommand{\R}{\mathbb{R}}		%

\DeclareMathOperator*{\argmax}{arg\,max}		%
\DeclareMathOperator*{\argmin}{arg\,min}		%

\DeclareMathOperator{\bigoh}{\mathcal{O}}		%

\newcommand{\nn}{\nonumber}		%

\newcommand{\eps}{\varepsilon}		%
\newcommand{\eg}{e.g.,\xspace}		%
\newcommand{\ie}{i.e.,\xspace}		%

\DeclareDocumentCommand\qst{s}{\IfBooleanTF{#1}{}{\quad}\text{s.t.}\quad}
\DeclareDocumentCommand\qand{s}{\IfBooleanTF{#1}{}{\quad}\text{and}\quad}
\DeclareDocumentCommand\qor{s}{\IfBooleanTF{#1}{}{\quad}\text{or}\quad}
\DeclareDocumentCommand\qwith{s}{\IfBooleanTF{#1}{}{\quad}\text{with}\quad}

\newcommand{\dd}{\mathrm{d}}

\newcommand{\obj}{F}
\newcommand{\ddv}[3]{{#1}'(#2; #3)} %

\newcommand{\dom}{\mathrm{dom}}
\newcommand{\grad}{\nabla}
\newcommand{\lap}{\Delta}

\newcommand{\meas}{\mathcal{M}}
\newcommand{\prob}{\mathcal{P}}

\newcommand{\Xsp}{\mathcal{X}} %
\newcommand{\Ysp}{\mathcal{Y}} %

\providecommand\from{}		%
\DeclarePairedDelimiterXPP{\KL}[1]{H}{(}{)}{}{%
\renewcommand\from{\nonscript\,\delimsize\|\nonscript\,\mathopen{}} #1}

\newcommand{\bregpot}{\varphi}
\DeclarePairedDelimiterXPP{\breg}[1]{D_\bregpot}{(}{)}{}{%
\renewcommand\from{\nonscript\,\delimsize\|\nonscript\,\mathopen{}} #1}

\DeclarePairedDelimiterXPP{\bregst}[1]{D_{\bregpot^*}}{(}{)}{}{%
\renewcommand\from{\nonscript\,\delimsize\|\nonscript\,\mathopen{}} #1}

\newcommand{\ex}{\mathbb{E}}

\providecommand\given{}		%
\DeclarePairedDelimiterXPP{\exof}[1]{\ex}{[}{]}{}{%
\renewcommand\given{\nonscript\,\delimsize\vert\nonscript\,\mathopen{}} #1}

\newcommand{\margx}[1]{{#1}_{\Xsp}}
\newcommand{\margy}[1]{{#1}_{\Ysp}}

\renewcommand{\P}{\mathsf{P}}
\renewcommand{\Q}{\mathsf{Q}}

\newcommand{\ddt}{\frac{\dd}{\dd t}}
\newcommand{\dds}{\frac{\dd}{\dd s}}
\newcommand{\half}{\nicefrac{1}{2}}

\newcommand{\convexind}{\mathrm{I}}

\newacro{SB}{Schr\"odinger bridge}
\newacro{IPF}{Iterative Proportional Fitting}
\newacro{OT}{optimal transport}
\newacro{MD}{mirror descent}
\newacro{SDE}{stochastic differential equation}
\newacro{NN}{neural network}
\newacro{APT}{asymptotic pseudo-trajectory}
\newacro{ICT}{internally chain-transitive}

\newcommand{\sinkeps}{Sinkhorn$_\eps$\xspace}
\newcommand{\stepsinkeps}{$\gamma$-Sinkhorn\xspace}
\newcommand{\schroeps}{Schr\"odinger$_\eps$\xspace}

\newcommand{\refmeas}{\pi^{\mathrm{ref}}_\eps}
\newcommand{\Sin}[1]{\mathrm{S}_{#1}}

\newcommand{\rnder}[2]{\frac{\dd #1}{\dd #2}}

\newcommand{\getbackcoup}[1]{\mathrm{Cpl}_{#1}}

\newcommand{\dualsp}{\mathcal{D}}

\newcommand{\opt}[1]{{#1}^{\mathrm{opt}}}
\newcommand{\revbm}{W^{\mathrm{R}}}

\newcommand{\lr}{\ell} %

\newcommand{\coupmu}{\crl*{\pi : \margx\pi = \mu}}

\newcommand{\cst}{\mathcal{C}}

\newcommand{\newmacro}[2]{\newcommand{#1}{\debug{#2}}}		%
\newcommand{\newop}[2]{\DeclareMathOperator{#1}{\debug{#2}}}		%

\newcommand{\debug}[1]{#1}		%

\newmacro{\start}{1}		%
\newmacro{\runalt}{k}		%
\newmacro{\run}{n}		%

\newmacro{\state}{g}		%
\newmacro{\dstate}{Y}		%

\newcommand{\curr}[1][\state]{\debug{#1}^{\run}}		%
\newcommand{\next}[1][\state]{\debug{#1}^{\run+1}}		%

\newmacro{\ctime}{t}		%
\newmacro{\ctimealt}{s}		%
\newmacro{\cstart}{0}		%

\newmacro{\horizon}{T}		%

\newmacro{\step}{\gamma}		%
\newmacro{\temp}{\eta}		%

\newmacro{\efftime}{\tau}		%
\newmacro{\tinv}{M}		%
\newcommand{\apt}[2][]{\state^{#1}(#2)}		%

\newop{\orcl}{\mathsf{V}}		%
\newop{\err}{\mathsf{U}}		%

\newmacro{\signal}{V}		%
\newmacro{\error}{W}		%
\newmacro{\brown}{W}		%

\newmacro{\bbound}{B}		%

\newmacro{\totbound}{S}		%
\newmacro{\noisepar}{\sdev}		%
\newmacro{\noisevar}{\variance}		%

\DeclarePairedDelimiterX{\setdef}[2]{\{}{\}}{#1:#2}		%

\begin{document}

\title{Sinkhorn Flow: A Continuous-Time Framework for\\ Understanding and Generalizing the Sinkhorn Algorithm}
\author{
  Mohammad Reza Karimi\thanks{Equal contribution. Corresponding email address: \texttt{mkarimi@inf.ethz.ch}}
\and 
Ya-Ping Hsieh\footnotemark[1]
\and
Andreas Krause
}
\date{ETH Z\"urich}

\maketitle

\begin{abstract}

Many problems in machine learning can be formulated as solving entropy-regularized optimal transport
on the space of probability measures. The canonical approach involves the \emph{Sinkhorn iterates},
renowned for their rich mathematical properties. Recently, the Sinkhorn algorithm has been recast
within the \emph{mirror descent} framework, thus benefiting from classical optimization theory
insights. Here, we build upon this result by introducing a \emph{continuous-time} analogue of the
Sinkhorn algorithm. This perspective allows us to derive novel variants of Sinkhorn schemes that are
robust to noise and bias. 
Moreover, our continuous-time dynamics not only generalize but also offer a unified perspective on several recently discovered dynamics in machine learning and mathematics, such as the ``Wasserstein mirror flow'' of \cite{deb2023wasserstein} or the ``mean-field Schr\"odinger equation'' of \cite{claisse2023mean}.

\end{abstract}

\section{Introduction}
\label{sec:introduction}

Many modern machine learning tasks can be reframed as solving an entropy-regularized \ac{OT} problem
over the space of {probability measures}. One particularly noteworthy instance that has attracted
significant attention is the \acdef{SB} problem, whose primary objective is to dynamically transform a given
measure into another measure. Consequently, \ac{SB} has found widespread application in diverse
domains that require an understanding of complex continuous-time systems, spanning applications such
as sampling \citep{bernton2019schr, huang2021schrodinger}, generative modeling
\citep{chen2021likelihood,de2021diffusion,pmlr-v139-wang21l}, molecular biology
\citep{holdijk2022path}, single-cell dynamics \citep{somnath2023aligned, pariset2023unbalanced,
bunne2023schrodinger}, and mean-field games \citep{liu2022deep}.

The canonical approach for solving entropy-regularized \ac{OT} is the \emph{Sinkhorn}
algorithm \citep{sinkhorn1967concerning} or the closely related \acdef{IPF} procedure
\citep{fortet1940resolution, kullback1968probability, chen2021optimal}. Traditionally, these
algorithms have been framed as an \emph{alternating projection} procedure and extensively
studied in this regard \citep{cuturi2013sinkhorn, peyre2019computational, ghosal2022convergence,
chen2016entropic}. Conversely, recent research  \citep{mishchenko2019sinkhorn, mensch2020online,
ballu2023mirror} sheds new light on Sinkhorn for discrete probability distributions by associating it
to the \acdef{MD} scheme \citep{nemirovsky1983problem, beck2003mirror}, thereby opening up new
avenues for understanding Sinkhorn through the lens of classical optimization theory. Furthermore,
\citet{aubin2022mirror, leger2021gradient} generalize this insight to the space of continuous
probability measures. To be more precise, these studies reveal that the Sinkhorn iterates can be
perceived as \ac{MD} steps, specifically implemented \emph{with step-size 1}.

Here, we advance this \acl{MD} perspective by introducing a novel, continuous-time variant of the
Sinkhorn algorithm on the space of probability measures. Our objectives are twofold. Firstly, we
deepen our comprehension of Sinkhorn iterates by emphasizing that the key components for
establishing convergence are not limited to the previous focus of the specific step-size 1, but are
inherently linked to the choice of the mirror map, objective function, and constraints. Building
upon this insight, we harness stochastic \ac{MD} analysis to yield novel Sinkhorn variations that,
unlike the traditional approach, maintain convergence even in the presence of noise and bias.
Secondly, we demonstrate that our continuous-time dynamics opens up 
a unified perspective on various intriguing dynamics recently explored in machine learning and
mathematics.

To summarize, we make the following contributions:
\begin{itemize}
\item We derive a specific formula for a discrete-time mirror descent scheme that encompasses
  Sinkhorn iterates as a special case by setting the step-size to 1. By driving the step-sizes to 0, we
  obtain a novel continuous-time version of the Sinkhorn iterates, comprising a pair of ``primal'' and
  ``dual'' systems referred to as the \sinkeps and \schroeps flows, respectively.

\item Importantly, both our novel discrete-time and continuous-time schemes preserve
  the essential properties of Sinkhorn needed for ensuring its convergence. This insight deepens our
  understanding by highlighting that Sinkhorn's convergence is primarily a result of the
  mirror descent algorithm with a specific mirror map, as opposed to the conventional interpretation
  centered around alternating projection. Leveraging this fresh perspective, we further devise new
  schemes accompanied by robust asymptotic and non-asymptotic guarantees.

\item With the aid of Otto calculus \citep{otto2001geometry}, we demonstrate that our dynamics
  formally encompass the ``Wasserstein Mirror Flow'' introduced by \cite{deb2023wasserstein} as well
  as the evolution proposed by \cite{claisse2023mean}. The former is motivated by modeling the
  behavior of Sinkhorn for \emph{unregularized} \acl{OT} (\ie when $\eps \to 0$ in
  \eqref{eq:eot-alt} below), while the latter seeks to relax the conditions for establishing the
  convergence of neural network training in the mean-field limit. Until now, these dynamics were
  treated in isolation, primarily due to the lack of a continuous-time mirror descent perspective.

\item We contextualize these findings in the \acl{SB} setting and provide a mirror descent
  interpretation of the \acl{IPF} procedure widely adopted in the machine learning community
  \citep{vargas2021solving, de2021diffusion,chen2021likelihood}. Additionally, we establish that
  these new mirror descent iterations can be expressed as \aclp{SDE} with an explicit drift formula.

\end{itemize}

\section{Background on Mirror Descent}
\label{sec:background}

In this section, we revisit the fundamental components of the classical mirror descent (MD) scheme
\citep{nemirovsky1983problem,beck2003mirror}.

\para{\ac{MD} as a ``minimizing movement'' scheme}
Let $\obj : \R^d \to \R$ be a differentiable objective function, $\bregpot : \R^d \to \R$ be
strictly convex and differentiable called the \emph{Bregman potential}, and $\cst \subseteq \R^d$ be a
convex constraint.
The {\ac{MD} algorithm} aims to find  $\min_{x \in \cst} F(x)$ by following the iterations
\begin{equation}\label{eq:mirror-descent-mms-when-grad-exists}
  x_{n+1} = \argmin_{x \in \cst} \crl*{ \inner{\grad \obj(x_n), x - x_n} 
            + \frac{\breg{x \from x_n}}{\gamma_n} }.
\end{equation}
Here, $\gamma_n$ is a sequence of step-sizes, and $\breg{x \from x_n}$ is the Bregman divergence
associated with $\bregpot$:
\begin{equation}\label{eq:bregman-divergence-def}
  \breg{x' \from x} \coloneqq \bregpot(x') - \bregpot(x) - \inner{\grad \bregpot(x), x' - x}.
\end{equation}
This is the \emph{minimizing movement} interpretation of \ac{MD}: At each iteration,
linearize the objective $\obj$ and minimize it while staying ``close'' to the previous iterate, where the measure of closeness is determined by the Bregman divergence.

\para{The dual perspective of \ac{MD}}

A particularly insightful approach for studying \ac{MD} employs the concept of \emph{convex duality}
\citep{rockafellar1997convex}. Recall that for a convex set $\cst$, the convex indicator function
$\convexind_\cst$ is defined as
$\convexind_\cst(x) = 0$ if $x\in \cst$ and $+\infty$ otherwise.
Let $\bregpot^*$ be the \emph{Fenchel conjugate} of $\bregpot + \convexind_\cst$:
\begin{equation}\label{eq:fenchel-pot-plus-indic}
  \begin{aligned}
    \bregpot^*(y) &= \sup_{x\in\Xsp} \crl{\inner{x,y} - (\bregpot + \convexind_\cst)(x)} \\
    &= \sup_{x\in \cst} \crl{\inner{x,y} - \bregpot(x)}.
  \end{aligned}
\end{equation}
As $\bregpot + \convexind_\cst$ is strictly convex, it holds that $\bregpot^*$ is essentially differentiable \citep{rockafellar1997convex}, and the Danskin's theorem implies
\[
  \grad \bregpot^* (y) = \argmax_{x\in \cst} \crl*{ \inner{x, y} - \bregpot(x)}.
\]
Thus, $\grad \bregpot^* (\grad \bregpot(x')) = \argmin_{x\in \cst} \breg{x \from x'}$. In particular,
for all $x \in \cst$, $\grad \bregpot^* (\grad \bregpot(x)) = x$. 

We shall call any $x \in \cst$ a \emph{primal point} and any $y \in \dom (\grad
\bregpot^*)$ a \emph{dual point}. Notice that a dual point $y$ uniquely identifies a primal point $x
= \grad \bregpot^*(y)$, but several dual points might correspond to the same primal point $x$, one
of which is $\grad\bregpot(x)$.%
\footnote{For a fixed primal point $\bar{x}\in\cst$, any $x\in\R^d$ such that $\argmin_{x' \in \cst} \breg{x'
  \from x} = \bar{x}$ satisfies $\grad\bregpot^*(\grad \bregpot(x)) = \bar{x}$.}

Now, let $y_0 = \grad\bregpot(x_0)$ and consider the following dual
iterations:
\begin{equation}\label{eq:mirror-descent-dual-iteration}
  \begin{cases}
    y_{n+1} = y_n - \gamma_n \grad \obj(x_n),\\
    x_{n+1} = \grad \bregpot^*(y_{n+1}),
  \end{cases}
\end{equation}
which can also be written solely in terms of the dual variables as
\[
  y_{n+1} = y_n - \gamma_n \grad \obj(\grad \bregpot^*(y_n)).
\]
Under certain mild conditions which hold in our setting, \eqref{eq:mirror-descent-dual-iteration}
coincides with \eqref{eq:mirror-descent-mms-when-grad-exists} \citep{beck2003mirror}.
Driving the step-size $\gamma_n\to 0$, we obtain the \emph{continuous-time} limit of the
dual iterations \citep{krichene2015accelerated, tzen2023variational}, called the \emph{mirror flow}:
\begin{equation}\label{eq:mirror-flow}
  \begin{cases}
    \ddt{y}(t) = -\grad \obj(x(t)),\\
    x(t) = \grad \bregpot^*(y(t)),
  \end{cases}
\end{equation}
Alternatively, we can write the mirror flow as
\begin{equation}\label{eq:mirror-flow-alt}
  \ddt{y}(t) = -(\grad \obj \circ \grad \bregpot^*)(y(t)).
\end{equation}

\para{MD on the space of probability measures}

While the discussion has thus far been limited to the Euclidean spaces, the extension of \ac{MD} to
the infinite-dimensional space of \emph{probability measures} presents no conceptual difficulty: One
can simply replace the inner product in \eqref{eq:bregman-divergence-def} with the \emph{duality
pairing} $\inner{ p ,f} \coloneqq \ex_p f$ \citep{halmos2013measure}, and the gradient operator with
the \emph{first variation} as defined in \cref{app:background}
\citep{hsieh2019finding,bauschke2003bregman,leger2021gradient}.

\section{Continuous-Time Sinkhorn Flows}
\label{sec:sinkhornflows}

This section constitutes our core contribution: The introduction of a pair of ``primal'' and
``dual'' dynamics, which can be viewed as the analogue of the Sinkhorn algorithm in continuous time.
Our derivation builds upon a well-established \ac{MD} interpretation of Sinkhorn, reviewed in detail
in Section \ref{sec:EOT-background}. In \cref{sec:sinkhorn-stepsizes}, we present a simple yet
pivotal extension of the \ac{MD} interpretation, serving as the foundation for our main result in
\cref{sec:sinknschro-flows}. We then showcase in \crefrange{sec:convergence}{sec:otto} how our novel
dynamics allows unifying existing dynamics as well as motivating new schemes with advantageous
properties compared to Sinkhorn. All proofs can be found in \cref{app:proofs-sinkhorn}.

\newcommand{\defeq}{\coloneqq}
\subsection{Background: Entropic Optimal Transport and the Sinkhorn Algorithm}
\label{sec:EOT-background}

We first recall the central properties of the entropy-regularized \ac{OT}; the materials are
classical, and presented \eg by \citet{peyre2019computational}.
Let $\mu$ and $\nu$ be two given probability measures on $\Xsp$ and $\Ysp$, respectively. Consider a
cost function $c: \Xsp \times \Ysp \to \R$ and a regularization parameter $\eps > 0$. 
The \emph{entropy-regularized \ac{OT}} is the minimization problem
\begin{equation}\label{eq:eot-alt}\tag{OT${}_\eps$}
  \mathrm{OT}_\eps(\mu, \nu) \defeq \min_{\pi \in \Gamma(\mu, \nu)} \ex_{\pi}[c] + \eps \KL{\pi \from \mu \otimes \nu}
\end{equation}
where $\Gamma(\mu, \nu)$ is the set of all couplings between $\mu$ and $\nu$, and $\KL{\cdot \from
\cdot}$ is the relative entropy.
One can alternatively rewrite \eqref{eq:eot-alt} as the \emph{static Schr\"odinger problem} (see
\cref{sec:SB} for an explanation of the terminology):
\begin{equation}\label{eq:eot}
  \min_{\pi \in \Gamma(\mu, \nu)} \KL{\pi \from \refmeas},
\end{equation}
where the reference measure $\refmeas$ is defined as 
$\dd\refmeas \propto \exp(-c/\eps)\,\dd(\mu\otimes \nu)$ and encodes all the information about
$\eps$ and the cost $c$. Without loss of generality, we assume that $c$ is normalized s.t.~$\dd\refmeas / \dd(\mu \otimes \nu) = \exp(-c/\eps)$.

The optimal solution $\opt\pi_\eps$ of
\eqref{eq:eot-alt} admits the following dual representation: There exists potential functions
$f:\Xsp\to\R$ and $g:\Ysp\to\R$, unique up to constants, such that
\begin{equation}\label{eq:schrodinger-pot}
  \dd\opt\pi_\eps = \exp\prn*{f\oplus g - \frac{c}{\eps}}\,\dd(\mu \otimes \nu) =  \exp(f\oplus g)\, \dd \refmeas.
\end{equation}
Here, we use the notation $(f \oplus g)(x, y) = f(x) + g(y)$, and call $f$ and $g$ the
\emph{Schr\"odinger potentials} of $\opt\pi_\eps$.%
\footnote{Some authors \citep[such as][]{nutzwiesel2022} prefer writing $\dd\opt\pi_\eps$ as
  $\exp\prn*{\frac{1}{\eps}(f \oplus g - c)}\,\dd(\mu \otimes \nu)$, and call these $f$
  and $g$ the Schr\"odinger potentials.}
Moreover, the reverse direction also holds: If a coupling $\pi \in \Gamma(\mu, \nu)$ has the form 
\eqref{eq:schrodinger-pot}, then it is the optimal solution of $\mathrm{OT}_\eps(\mu,\nu)$.

\para{The Sinkhorn Algorithm}
A popular method for solving \eqref{eq:eot-alt} is the Sinkhorn algorithm \citep{sinkhorn1967concerning, cuturi2013sinkhorn}: Starting from $\pi^0 \coloneqq \refmeas$, the algorithm iterates as
\begin{equation}\label{eq:sinkhorn-classic-update}\tag{Sink${}_1$}
  \begin{aligned}
    \pi^{n+\half} &\coloneqq \argmin \crl*{ \KL{\pi \from \pi^{n}} : \margy\pi = \nu },\\
    \pi^{n+1} &\coloneqq \argmin \crl{ \KL{\pi \from \pi^{n+\half}} : \margx\pi = \mu }.
  \end{aligned}
\end{equation}
Here, $\margx\pi$ denotes the $\Xsp$-marginal of $\pi$.
We use the notation $\pi^{n+1} = \Sin{1}[\pi^{n}]$ to represent a full
\eqref{eq:sinkhorn-classic-update} iteration. A key attribute of the Sinkhorn algorithm is
that all the information concerning the cost $c$ and the regularization parameter $\eps$ is embedded in the
initialization of $\pi^0$; the operator $\Sin{1}$ itself is independent of $c$ and $\eps$.

The special structure of algorithm \eqref{eq:sinkhorn-classic-update} guarantees that the
iterations $\pi^n$ admit the form $\exp\prn*{f^n_\eps \oplus g^n_\eps}\,\dd\refmeas$ for
some potentials $f^n_\eps$ and $g^n_\eps$. Furthermore, since the $\Xsp$-marginal of each successive
$\pi^{n}$ is always $\mu$, we can determine $f^n_\eps$ from $g^n_\eps$ as:
\begin{equation}\label{eq:get-f-from-g}
  f^n_\eps(x) = -\log \int \exp \prn*{ g^n_\eps(y) - \frac{c(x,y)}{\eps}}\,\nu(\dd y).
\end{equation}
This also implies that $\pi^n$ can be recovered solely from $g^n_\eps$. We articulate this connection by
defining the operator $\getbackcoup{\eps}$ as $\pi^n =
\getbackcoup{\eps}[g^n_\eps]$.

\para{Mirror Descent interpretation}
It is recently shown in \citep{aubin2022mirror} that the Sinkhorn algorithm can be viewed as an
\ac{MD} iteration in the space of probability measures. Specifically, by defining the objective
function $\obj(\pi) \defeq \KL{\margy\pi
\from \nu}$, the Bregman potential $\bregpot(\pi) \defeq \KL{\pi \from \refmeas}$,
and the constraint set $\cst \defeq \crl*{ \pi : \margx\pi = \mu }$, the Sinkhorn update
\eqref{eq:sinkhorn-classic-update} corresponds to the \ac{MD} update
\begin{equation}\label{eq:sinkhorn-classic-md}
  \pi^{n+1} = \argmin_{\pi \in \cst} \crl*{ \inner{\delta\obj(\pi^{n}), \pi - \pi^{n}} + 
  \breg{\pi \from \pi^{n}}},
\end{equation}
where $\delta\obj(\pi)$ is the first variation of $\obj$ at $\pi$ (see \cref{app:background} for the
related background).

\subsection{Sinkhorn with {\em Arbitrary} Step-sizes}
\label{sec:sinkhorn-stepsizes}
Comparing \eqref{eq:sinkhorn-classic-md} with the \ac{MD} update rule 
\eqref{eq:mirror-descent-mms-when-grad-exists}, we see that the classical Sinkhorn
corresponds to \ac{MD} \emph{with constant step-size 1}. However, once the connection to \ac{MD} is
established, we can use arbitrary step-sizes $\gamma_n$ to get the \stepsinkeps iteration, defined
as follows:
\begin{definition}[\stepsinkeps iteration]\label{def:sinkhorn-step-md}
  Let $\obj(\pi) = \KL{\margy\pi \from \nu}$, $\bregpot(\pi) = \KL{\pi \from \refmeas}$, and $\cst =
  \coupmu$.
  Starting from $\pi^0_\gamma \coloneqq \refmeas$, we define the iterates $\pi^n_\gamma$ as
  \[
    \pi^{n+1}_{\gamma} = \argmin_{\pi \in \cst} 
    \crl*{ \inner{\delta\obj(\pi^{n}_{\gamma}),\pi - \pi^{n}_{\gamma}} + 
        \frac{\breg{\pi \from \pi^{n}_{\gamma}}}{\gamma_n}}
  \]
  and write $\pi^{n+1}_{\gamma} = \Sin{\gamma_n}[\pi^{n}_{\gamma}]$.
\end{definition}

The following lemma is a simple observation that lies at the heart of all forthcoming derivations:
\begin{restatable}{lemma}{closedformstepsinkhorn}\label{lem:closed-form-step-sinkhorn}
  The \stepsinkeps iterates $\pi^n_\gamma$ defined as in \cref{def:sinkhorn-step-md} correspond to the update rule:
  \begin{align}
      \pi^{n+\half}_\gamma &\coloneqq \argmin \crl{ \KL{\pi \from \pi^{n}_\gamma} : \margy\pi = \nu}, \nn \\
      \pi^{n+1}_\gamma     &\coloneqq \argmin \crl{ \gamma_n \KL{\pi \from \pi^{n+\half}_\gamma} \label{eq:sinkhorn-step-update}\tag{Sink${}_\gamma$}
      + (1-\gamma_n) \KL{\pi \from \pi^{n}_\gamma} : \margx\pi = \mu }.
  \end{align}
\end{restatable}
Observe that in \eqref{eq:sinkhorn-step-update} the half-iteration updates %
are exactly the same as in the classical Sinkhorn \eqref{eq:sinkhorn-classic-update}. However,
unlike \eqref{eq:sinkhorn-classic-update}, integer iterations $\pi^{n+1}_\gamma$'s are now
computed based on both \smash{$\pi^{n+\half}_\gamma$} \emph{and} $\pi^{n}_\gamma$, hence losing the
interpretation of being a KL-projection step from \smash{$\pi^{n+\half}_\gamma$}. Nevertheless, the
next lemma establishes that \emph{the existence of Schr\"odinger potentials is retained} for
\eqref{eq:sinkhorn-step-update}:
\begin{restatable}{lemma}{formulaKantorovicStepSinkhorn}
  \label{lem:formula-kantorovic-step-sinkhorn}
  The \stepsinkeps iterates $\pi^n_\gamma$ in \eqref{eq:sinkhorn-step-update} admit the
  representation
  \begin{equation}
  \label{eq:gamma-sink-potentials}
    \dd\pi^n_\gamma = \exp\prn*{f^n_{\gamma} \oplus g^n_{\gamma}}
    \,\dd\refmeas.
  \end{equation}
  Moreover, the potentials $g^n_{\gamma}$ satisfy the recursion
  \[
    g^{n + 1}_{\gamma} = g^n_{\gamma} - \gamma_n \log \rnder{\margy{(\pi^n_\gamma)}}{\nu},
  \]
  and $f^{n+1}_{\gamma}$ is computed from $g^{n+1}_\gamma$ as in \eqref{eq:get-f-from-g}.
\end{restatable}

In \cref{sec:convergence}, we will establish the convergence of the generalized iterates in
\eqref{eq:sinkhorn-step-update}. In essence, this means that the dual representation provided in
\eqref{eq:gamma-sink-potentials}, which arises from the choice of the objective function $\obj$, the
Bregman potential $\bregpot$, and the constraint set $\cst$ in \cref{def:sinkhorn-step-md}, is
sufficient for ensuring convergence without resorting to the conventional alternating projection
interpretation.

\begin{remark*}
After the submission of our work, a preprint authored by \citet{chopin2023connection} was made available on arXiv. In the context of the \emph{tempering} for Monte Carlo methods, the authors derive a formula that bears a resemblance to our \eqref{eq:sinkhorn-step-update}. It is worth noting, however, that their framework does not handle the constraint $\cst$, which plays a pivotal role in the updates of potentials \eqref{eq:gamma-sink-potentials}. Since the dynamics of potentials are at the heart of our subsequent continuous-time flows and extensions for the \acl{SB}, our work and that of  \citet{chopin2023connection} lead to somewhat divergent developments: Our contributions primarily revolve around the generalization and unification of Sinkhorn schemes, while their work is more focused on the development of provable tempering schedules. %
\end{remark*}

\subsection{\sinkeps and \schroeps Flows}
\label{sec:sinknschro-flows}

Analyzing an algorithm's continuous-time limits often provides a more manageable analytical
perspective than for its discrete-time counterpart. Further, continuous-time dynamics are beneficial
for implementing \emph{stochastic approximation} techniques (see \cref{sec:convergence}), which are
essential for determining the algorithm's convergence. The following proposition characterizes the limiting behavior of operator $\Sin{\gamma}$ as $\gamma \to 0$.

\begin{restatable}{proposition}{timeDerivativeSinkhorn}
  \label{prop:time-derivative-sinkhorn}
  Fix a coupling $\pi\in\coupmu$. For any $\gamma > 0$, let $\pi^\gamma =
  \Sin{\gamma}[\pi]$. Then, 
  \[
    \frac{\dd}{\dd\gamma}\Big\lvert_{\gamma=0}
    \log \pi^\gamma(x,y) = -\log \frac{\dd\margy\pi}{\dd\nu}(y)
    + \ex_{\pi(\cdot \mid x)}\brk*{\log \frac{\dd\margy\pi}{\dd\nu}}.
  \]
  Moreover, if $\dd\pi = \exp(f \oplus g)\,\dd\refmeas$, then for all $\gamma>0$,  
  $\dd\pi^\gamma = \exp(f^\gamma \oplus g^\gamma)\,d\refmeas$, and $f^\gamma$ and $g^\gamma$ satisfy
  \begin{align*}
    \frac{\dd}{\dd\gamma}\Big\lvert_{\gamma = 0} g^\gamma(y)
      &= -\log \frac{\dd\margy\pi}{\dd\nu}(y), \\
    \frac{\dd}{\dd\gamma}\Big\lvert_{\gamma = 0} f^\gamma(x) 
      &= \ex_{\pi(\cdot\mid x)}\brk*{\log\frac{\dd \margy\pi}{\dd \nu}}.
  \end{align*}
\end{restatable}
In view of \cref{prop:time-derivative-sinkhorn}, we are now ready to define the \emph{\sinkeps} and the \emph{\schroeps flows}.
\begin{definition}\label{def:sink-flow-schro-flow}
Consider the set of all joint distributions on $\Xsp \times \Ysp$ that solve \eqref{eq:eot-alt} with cost function $c$ for their own marginals:
\begin{equation}
\label{eq:def-Pi_c-eps}
\Pi_{c, \eps} \defeq  \crl*{ \pi : \pi \text{ solves } \mathrm{OT}_\eps(\pi_{\Xsp}, \pi_{\Ysp}) }.
\end{equation}
For any $\pi^0_\eps \in \coupmu \cap \Pi_{c,\eps}$, we
construct a curve $(\pi^t_\eps)_{t\geq 0}$ whose velocity is determined by 
  \begin{equation}\label{eq:sink-logpi-flow}
    \begin{aligned}
      \ddt \log \rnder{\pi^t_\eps}{\refmeas} (x,y) &= -\log \frac{\dd\margy{(\pi^t_\eps)}}{\dd\nu}(y)+ \ex_{\pi^t_\eps(\cdot \mid x)}\brk*{\log
      \frac{\dd\margy{(\pi^t_\eps)}}{\dd\nu}}.
    \end{aligned}
  \end{equation}
  We call the mapping $(\pi^0_\eps, t) \mapsto \pi^t_\eps$ the \emph{\sinkeps flow}.
  Similarly, we call the mapping $(g^0_\eps, t) \mapsto g^t_\eps$ the \emph{\schroeps flow}, which describes the evolution of the
  Schr\"odinger potential corresponding to $\pi^t_\eps$:
  \begin{equation}\label{eq:sink-schrodinger-pot-flow}
    \ddt g^t_\eps = -\log \frac{\dd\margy{(\pi^t_\eps)}}{\dd\nu} 
    = -{\delta F}(\pi^t_\eps). %
  \end{equation}
\end{definition}
\begin{remark*}
Using $\pi^t_\eps = \getbackcoup{\eps}[g^t_\eps]$, the \schroeps flow
\eqref{eq:sink-schrodinger-pot-flow} can be written solely in terms of $g^t_\eps$. The other
direction also holds: if we integrate \eqref{eq:sink-schrodinger-pot-flow} to get $g^t_\eps$, we can
recover the \sinkeps flow \eqref{eq:sink-logpi-flow}. In short, the two flows
\eqref{eq:sink-logpi-flow} and \eqref{eq:sink-schrodinger-pot-flow} are equivalent.
\end{remark*}

\para{Mirror Flow interpretation}
\newcommand{\primalsp}{\mathcal{P}}

As the \sinkeps and \schroeps flows emerge from driving the step-size of an \ac{MD} iterate to zero,
it is natural to anticipate a \emph{mirror flow} interpretation in the sense of
\eqref{eq:mirror-flow}. In this section, we make this link precise.

Consider the primal space $\primalsp$ to be the space of probability measures over $\Xsp \times
\Ysp$ having smooth densities with respect to the Lebesgue measure, and the dual space $\dualsp
\defeq L^1(\Xsp\times \Ysp)$ to be the space of integrable functions.
The first variation of the Bregman potential $\bregpot(\pi) = \KL{\pi \from \refmeas}$ gives a link
from $\primalsp$ to $\dualsp$:
\begin{equation}
  \delta \bregpot(\pi) = \log  \rnder{\pi}{\refmeas} \in \dualsp, \quad
  \forall\pi\in\primalsp,\, \pi \ll \refmeas.
\end{equation}
Moreover, the Fenchel conjugate $\bregpot^*$ of $\bregpot + \convexind_\cst$, defined in
\eqref{eq:fenchel-pot-plus-indic}, gives a link from $\dualsp$ to $\cst = \coupmu  \subset \primalsp$ through its
first variation. 

Crucially, unlike prior studies such as
\citep{aubin-frankowskietal2022,ballu2023mirror,mishchenko2019sinkhorn}, we incorporate the
$\convexind_\cst$ in our definition of $\bregpot^*$. This plays a vital role in preventing the
emergence of a differential \emph{inclusion}, as opposed to an equation, in the dual iteration,
which is pivotal in our interpretation of mirror flow and the Otto calculus.

\begin{restatable}{lemma}{formulafenchelconjugaterelentconstrained}
  \label{lem:formula-fenchel-conjugate-relent-constrained}
  The Fenchel conjugate $\bregpot^*$ of $\bregpot + \convexind_\cst$ evaluated at $h \in \dualsp$ is
  given by $\bregpot^*(h) = \inner{\hat{\pi}, h} - \KL{\hat{\pi} \from \refmeas}$, where
  \begin{equation}
  \label{eq:extremizer}
    \hat{\pi}(x,y) \defeq \frac{\refmeas(x,y)\, e^{h(x,y)}}{\int \refmeas(x,y')\, e^{h(x,y')}\,dy'} \, \mu(x)
    \in \cst.
  \end{equation}
  Moreover, one has $\delta \bregpot^*(h) = \hat{\pi}$ defined in \eqref{eq:extremizer}.
\end{restatable}
\newcommand{\primalvar}{\hat{\pi}}
\newcommand{\dualvar}{h}

Using these formulas for $\delta \bregpot$ and $\delta \bregpot^*$, we can then define an
infinite-dimensional \emph{mirror flow} as follows. Fix $\primalvar^0_\eps \in \Pi_{c,\eps} \cap
\coupmu$ and $\dualvar^0_\eps \defeq \delta\bregpot(\primalvar^0_\eps) \in \dualsp$,  and consider%
\begin{equation}\label{eq:mirror-flow-measures}
  \begin{cases}
    \ddt \dualvar^t_\eps = -\delta \obj(\primalvar^t_\eps),\\
    \primalvar^t_\eps = \delta \bregpot^*(\dualvar^t_\eps),
  \end{cases}%
  \end{equation}which can be equivalently written as
\begin{equation}\label{eq:mirror-flow-alt-measures}
  \ddt \dualvar^t_\eps  = -(\delta \obj \circ \delta \bregpot^*)(\dualvar^t_\eps).
\end{equation}

We then have:
\begin{restatable}{thm}{mflow}
\label{thm:mflow}
The dynamics \eqref{eq:mirror-flow-measures} or \eqref{eq:mirror-flow-alt-measures} coincide with
the \schroeps flow \eqref{eq:sink-schrodinger-pot-flow}, and the corresponding $(\primalvar^t_\eps)_{t\geq 0}$ solves the \sinkeps flow \eqref{eq:sink-logpi-flow} starting at
$\primalvar^0_\eps$.
\end{restatable}

Comparing \crefrange{eq:mirror-flow-measures}{eq:mirror-flow-alt-measures} with
\crefrange{eq:mirror-flow}{eq:mirror-flow-alt}, we thus see that \sinkeps and \schroeps flows can be
seen as the analogue of the mirror flow in the space of probability measures. 

\subsection{Convergence of \sinkeps Flow and $\gamma$-Sinkhorn Iterates}
\label{sec:convergence}

The flexibility of the variable step-sizes in \stepsinkeps schemes offers a
straightforward framework for improving the traditional Sinkhorn algorithm. Furthermore, the
continuous-time \sinkeps and \schroeps flows pave the way for integrating the powerful machinery
of \emph{stochastic approximation} techniques
\citep{mertikopoulos2023unified,karimi2022dynamical,karimi2022dynamics}. In this section, we
illustrate how to leverage our theory to enhance the convergence of Sinkhorn schemes in scenarios
involving \emph{noisy gradients}.

\para{Rate analysis for the \sinkeps flow}
Before we proceed to present improved Sinkhorn schemes, we first establish the convergence rate of the continuous-time \sinkeps flow:
\begin{restatable}{thm}{sinkepsrate}
\label{thm:sinkepsrate}
  Starting from $\pi^0_\eps \in \Pi_{c,\eps} \cap \coupmu$, consider the \sinkeps flow $\pi^t_\eps$
  and the corresponding \schroeps flow $g^t_\eps$. Then, 
  \[
    \obj(\pi^t_\eps) \leq \frac{ \bregst{ g^0_\eps \from \opt g_\eps }}{t} = \bigoh\prn*{t^{-1}},
  \]
  where $\opt g_\eps$ is the Schr\"odinger potential of the optimal coupling for
  \eqref{eq:eot-alt}.  That is, the $\Ysp$-marginal of $\pi^t_\eps$ converges (in relative entropy)
  to $\nu$ with the rate $1/t$.
\end{restatable}
While our proof of \cref{thm:sinkepsrate} is guided by the mirror flow
formalism presented in the previous section, it does \emph{not} follow directly from existing results for
mirror flows such as \citep{krichene2015accelerated, tzen2023variational}. This is primarily due to
the presence of the additional constraint $\cst$, which is absent in conventional mirror descent
analyses.

\para{Convergent Sinkhorn under noise}

In entropy-regularized \ac{OT}, \acp{NN} are commonly used to parameterize the transport plans.
Typically, the Sinkhorn iterations \eqref{eq:sinkhorn-classic-update} are employed, requiring
solving an infinite-dimensional optimization problem, approximated via multiple stochastic gradient
steps over \acp{NN}. However, inherent stochasticity in computations can \emph{prevent convergence}
when $\delta \obj$ in \eqref{eq:sinkhorn-classic-md} is replaced by a noisy estimate
$\tilde{\delta}\obj$, necessitating a remedy \citep{hanzely2021fastest}.

In this section, we introduce two improvements. First,  \cref{thm:gammaconverg} shows that using
our variable step-size method \eqref{eq:sinkhorn-step-update} with $\gamma_n =
\bigoh\prn{n^{-1/2}}$, one maintains a convergence rate of $\bigoh\prn{n^{-1/2}}$, when the
``stochastic gradients'' remain unbiased with finite variance. Second, \cref{thm:sinkapt} establishes
asymptotic \emph{last-iterate} convergence when one employs stochastic \emph{and} biased gradient
estimates.

\begin{restatable}{thm}{gammaconverg}
\label{thm:gammaconverg}
Suppose that we have a stochastic estimate $\tilde{\delta}\obj$ of ${\delta}\obj$ such that $\ex
\brk{\tilde{\delta}\obj(\pi)} = {\delta}\obj(\pi)$ and $\ex \brk{
\Nrm{\tilde{\delta}\obj(\pi)}_{\infty}^2 } \leq \sigma^2 < \infty$ for all $\pi$. Consider the
iterations $\pi^n_\gamma$ generated by \eqref{eq:sinkhorn-step-update} using $\tilde{\delta}\obj$ 
and a fixed step-size $\gamma$. Then we have, with
$\bar{\pi}_\gamma^n \defeq \frac{1}{n} \sum_{k=0}^{n}  {\pi}_\gamma^k $,
\begin{equation}
\label{eq:gamma-rate}
\ex \brk*{  \KL{(\bar{\pi}_\gamma^n)_\Ysp \from \nu }  } \leq \frac{\KL{ \opt{\pi}  \from  \refmeas}}{\gamma n} + \gamma \sigma^2.
\end{equation}
\end{restatable}

The proof of \cref{thm:gammaconverg} is established by combining our framework with a classical analysis of stochastic Bregman schemes \citep{dragomir2021fast,hanzely2021fastest}.

\newcommand{\bias}{\lambda}
\newcommand{\noise}{\omega}

\newcommand{\filter}{\mathcal{F}}

While \eqref{eq:gamma-rate} immediately yields an $\bigoh(n^{-1/2})$ convergence rate, there are two significant drawbacks in \cref{thm:gammaconverg}. First, since the stochastic estimate $\tilde{\delta}\obj$ aims to capture noise introduced during the intermediate optimization procedures for \acp{NN}, the unbiasedness assumption is rather restrictive. Second, even if $\tilde{\delta}\obj$ is unbiased, we are still required to produce an \emph{ergodic} iterate $\bar{\pi}_\gamma^n$, whereas in practice, the last iterate ${\pi}_\gamma^n$ is often the most utilized. To address these issues, we leverage stochastic approximation analysis, which relies on the \emph{continuous-time} convergence in \cref{thm:sinkepsrate} to prove the following (see \cref{thm:sinkapt-full} for details):
\begin{thm}[Informal] \label{thm:sinkapt}
  Let $\pi^n$ be the sequence of measures generated by \eqref{eq:sinkhorn-step-update} using
  noisy and biased gradients $\tilde{\delta}F$, along with a step-size rule $\gamma_n$ such that $\sum\gamma_n
  = \infty$ and $\sum\gamma^2_n < \infty$. 
  Then $\lim_{n\to \infty}\pi^n = \opt{\pi}_\eps$ almost surely, if the biases vanish asympotically
  and the noises have uniformly bounded variance.
\end{thm}
\cref{thm:sinkapt} offers two advantages over \cref{thm:gammaconverg}. First, it replaces ergodic
convergence with the more desirable \emph{last-iterate} convergence. Secondly, if we consider
the bias as the error during the optimization of the \ac{NN} at each step of
\eqref{eq:sinkhorn-step-update}, then \cref{thm:sinkapt} allows for a level of flexibility where
the precision of the intermediate steps may progressively improve, instead of always requiring
perfect optimization as stipulated by the unbiased assumption in \cref{thm:gammaconverg}. However,
we acknowledge that these advantages come at the cost of losing a non-asymptotic rate.

\newcommand{\dilatedg}{\tilde{g}}
\newcommand{\kotpot}{g_0}

\subsection{Otto Calculus for \sinkeps and \schroeps Flows}

\label{sec:otto}

The Otto calculus, a groundbreaking development in 21st-century mathematics, has found far-reaching implications in machine learning applications \citep{otto2001geometry,ambrosio2005gradient,villani2008optimal}. Leveraging this powerful
framework, we showcase how our \sinkeps and \schroeps flows offer a unified template for various
recently discovered dynamics that hold significant relevance to machine learning \citep{deb2023wasserstein,claisse2023mean}.

\para{Wasserstein Mirror Flow}
Consider the time-dilated Schr\"odinger potential as $\dilatedg^t_\eps \defeq g^{t / \eps}_\eps$.
Recall that $\dilatedg^t_\eps \equiv \dilatedg^t_\eps(y)$ is a function on $\Ysp$. By taking the gradient with respect to the $y$ variable in \eqref{eq:sink-schrodinger-pot-flow}, we obtain:
\begin{align}
\label{eq:dilatedg-dynamics}
  \ddt \grad (\eps \dilatedg^t_\eps) = \ddt \grad_{\mathbb{W}_2}\textup{OT}_\eps(\mu, \cdot) = -\grad_{\mathbb{W}_2}\obj( \pi^t_\eps),
\end{align}
where $\grad_{\mathbb{W}_2} \obj(\pi) \defeq \grad \delta \obj(\pi)$ is the \emph{Wasserstein gradient} of $\obj$ \citep{otto2001geometry} and similar for $\grad_{\mathbb{W}_2} \textup{OT}_\eps$ \citep{benamouetal2023}, which is evaluated at $(\getbackcoup{\eps}[\tilde{g}^t_\eps])_\Ysp$. 

Now, let $c(x,y) = \frac{1}{2}\nrm*{x-y}^2$ in \eqref{eq:eot-alt}. In this case, it is well-known
that, as $\eps\to 0$, $\opt\pi_\eps$ converges to the optimal $\mathbb{W}_2$ coupling
(denoted by $\varpi$) and $\eps \opt{g}_\eps$ converges to the corresponding Kantorovich potential
(denoted by $\kotpot$) \citep{pooladian2021entropic, chiarini2023gradient}. Thus, by driving
$\eps\to 0$ in \eqref{eq:dilatedg-dynamics}, we formally get:
\begin{equation}
\label{eq:wasserstein-mirror-flow}
  \ddt \grad \kotpot^t = \ddt \grad_{\mathbb{W}_2} \mathrm{OT}_0(\mu, \margy\varpi^t) = -\grad_{\mathbb{W}_2}\obj(\varpi^t)
\end{equation}where $\varpi^t$ is the optimal $\mathbb{W}_2$ coupling of $\mu$ and
$\margy\varpi^t$, and $ \kotpot^t$ is the corresponding Kantorovich potential. This
equation is exactly the ``Wasserstein Mirror Flow'' of \citet[equation
(2.3)]{deb2023wasserstein} proposed to study Sinkhorn for the unregularized \ac{OT}. Therefore, the dynamics discovered by \citet{deb2023wasserstein} can be
seen as a limiting case of the more general \sinkeps and \schroeps flows.

\para{JKO Flow with Relative Entropy}
Recently, motivated by the mean-field limit of neural network training, \citet{claisse2023mean} study a variation of the JKO flow of a functional $\obj$
\citep{jordan1998variational}, replacing the $\mathbb{W}_2$ distance with the relative entropy:
\begin{equation}\label{eq:claisse-mms}
  p^{n+1}_\gamma \coloneqq \argmin_{p} \crl*{ \obj(p) + \gamma^{-1} \KL{ p \from p^{n}_\gamma }},
\end{equation}
where the minimization is over the set of all distributions with regular densities. In the limit $\gamma\to 0$, they
show that such as a scheme converges to the flow
\begin{equation}\label{eq:claisse-flow}
  \ddt \log p_t = -{\delta F}(p_t)
\end{equation}
which formally resembles \eqref{eq:sink-schrodinger-pot-flow}. However, there are two
important differences. 
Firstly, \eqref{eq:claisse-mms} constitutes an \emph{unconstrained} minimization problem, whereas
our mirror descent scheme seeks minimizers within the constraint set
$\cst = \coupmu$, giving rise to the central notion of Schr\"odinger potentials.
Secondly, the update \eqref{eq:claisse-mms} involves the original objective function $\obj$ in the
minimization, whereas our scheme employs the linearized objective. While the second difference
becomes negligible as $\gamma \to 0$ (intuitively, there would be no distinction between $\obj$
and its linearization), the impact of the constraint set $\cst$ persists. Consequently, one can view
\eqref{eq:claisse-flow} as an \emph{unconstrained} mirror flow for the objective $\obj$ with the
entropy as the Bregman potential.

\section{Extension to Schr\"odinger Bridges}
\label{sec:SB}

\newcommand{\refsde}{\P^{\textup{ref}}}
In \cref{sec:sinkhornflows}, we established the continuous-time variant of the Sinkhorn
iterates for \eqref{eq:eot-alt}, which pertains to the ``static'' entropy-regularized \ac{OT}. Motivated by the strong connections to diffusion models, in this section, we broaden our scope to encompass the \emph{dynamical} scenario, often referred to as
the \acdef{SB} problem. Beyond adapting the results in \cref{sec:sinkhornflows} to the
\ac{SB}, we provide additional insights by demonstrating that each time point in the continuous-time
\ac{SB} flow can be characterized as a \acli{SDE} with a well-defined drift formula. All proofs can be found in \cref{app:diffusion}.

\subsection{Review of Schr\"odinger Bridges}

\newcommand{\refdrift}{b^{\textup{ref}}}

Given two probability measures $\mu_0, \mu_T$ on $\R^d$, the \ac{SB} refers
to the following entropy minimization problem over the space of all \emph{stochastic processes} over $[0,T]$:
\begin{equation}\label{eq:schrodinger-problem}\tag{SB}
  \min_{\P} \crl*{ \KL{\P \from \refsde} : \P_0 = \mu_0,\,\P_T = \mu_T},
\end{equation}
where $\refsde$ is a given path measure induced by the solutions of the \acl{SDE}
\begin{equation}\label{eq:reference-sde}
  \dd X_t = \refdrift_t(X_t)\,\dd t + \sigma\dd W_t,%
\end{equation} 
and $\P_t$ is the marginal of $\P$ at time $t$. It turns out that solving
\eqref{eq:schrodinger-problem} is intimately related to solving the \emph{static} Schr\"odinger problem \citep{leonard2014}:
\begin{equation}\label{eq:SB_BM} %
  \min_{\pi \in \Gamma(\mu_0, \mu_T)} \KL{\pi \from \refsde_{0,T}}.
\end{equation}

\para{Connection to Entropy-Regularized OT}

In the case where $\refsde$ is the
law of a \emph{reversible} Brownian motion on $[0,1]$ with diffusion parameter $\sigma$
\citep{leonard2014}, the joint distribution of the end time points satisfies 
$\refsde_{0,1}(\dd x, \dd y) \propto \exp\prn{-\nrm{x-y}^2/2\sigma^2}$.
Therefore, \eqref{eq:SB_BM} becomes an instance of entropy-regularized \ac{OT} \eqref{eq:eot} with the cost $c(x,y) = \frac{1}{2}\nrm{x - y}^2$ and $\eps = \sigma^2$. Thus, \eqref{eq:schrodinger-problem} can be viewed as the \emph{dynamic} formulation of \eqref{eq:eot-alt} where,
instead of merely seeking an optimal coupling $\opt{\pi}_\eps$, one solves for an entire stochastic
process that transforms $\mu_0$ into $\mu_1$.

\para{\acl{IPF}}
The classical algorithm for solving \eqref{eq:schrodinger-problem} is the \acdef{IPF} procedure,
which can be seen as the \emph{dynamic} version of the Sinkhorn scheme: Starting from $\P^0 = \refsde$, define for $n\geq 0$,
\begin{equation}\label{eq:ipf}\tag{IPF}
  \begin{aligned}
    \P^{n + \half} &= \argmin \crl{ \KL{\P \from \P^{n}} : \P_T = \mu_T }, \\
    \P^{n + 1} &= \argmin \crl{ \KL{\P \from \P^{n+\half}} : \P_0 = \mu_0 }.
  \end{aligned}
\end{equation}
Notice that, in complete analogy to Sinkhorn, we have $\P^n_0 = \mu_0$ and \smash{$\P^{n+\half}_T =
\mu_T$}
for all $n\geq 0$.

With this background, we now present an approach to \ac{IPF} for \eqref{eq:schrodinger-problem} that parallels the Sinkhorn for \eqref{eq:eot-alt}.

\subsection{\ac{IPF} as Mirror Descent}

Similar to the Sinkhorn algorithm, we show that \ac{IPF} can be interpreted through the lens of
\ac{MD}. This finding serves as the dynamic counterpart to \citep[\textbf{Proposition
5}]{aubin2022mirror}.

\begin{restatable}{proposition}{ipfismd}
  \label{prop:ipf-is-md}
  The iterations $\P^n$ of \eqref{eq:ipf} satisfy 
  \begin{equation}\label{eq:mirror-descent-template}
    \P^{n+1} = \argmin_{\P \in \cst} \crl{ \inner{\delta\obj(\P^{n}), \P - \P^{n}} + \breg{\P \from \P^{n}} },
  \end{equation}
  with $\obj(\P) \defeq \KL{\P_T \from \mu_T}$, $\bregpot(\P) \defeq \KL{\P \from \refsde}$, and 
  $\cst \defeq \crl*{ \P : \P_0 = \mu_0}$.
\end{restatable}

In other words, \eqref{eq:ipf} is equivalent to an \ac{MD} iteration with step-size 1.

\para{$\gamma$-IPF iterations}

Upon recognizing that \ac{IPF} can be interpreted as \ac{MD} iterations with a step-size of 1, we can proceed to investigate the \ac{MD} iteration \eqref{eq:mirror-descent-template} with an arbitrary step-size $\gamma_n$:
\begin{equation}\label{eq:mirror-descent-stepsize-template}
  \P^{n+1} = \argmin_{\P \in \cst} \crl*{ \inner{\delta\obj(\P^{n}), \P - \P^{n}} + \frac{\breg{\P \from \P^{n}}}{\gamma_n} }.
\end{equation}
A similar calculation to that of \cref{lem:closed-form-step-sinkhorn} reveals that \eqref{eq:mirror-descent-stepsize-template} can be equivalently expressed as:
\begin{equation}
\tag{$\gamma$-IPF}
\label{eq:gamma-IPF}
  \P^{n+1} = \argmin_{\P \in \cst} \crl*{ \gamma_n\, \KL{\P \from \P^{n+\half}}
  + (1 - \gamma_n)\,\KL{\P \from \P^n}}.
\end{equation}

In analogy to the $\gamma$-Sinkhorn iterations, we call the update rule in \eqref{eq:gamma-IPF} the
\emph{$\gamma$-\ac{IPF} scheme}.

\subsection{The SDE Representation of $\gamma$-\ac{IPF}}

So far, we have shown that the results in \cref{sec:sinkhornflows} can be straightforwardly extended to \ac{SB} setting.
However, the true significance of the \ac{SB} formulation becomes apparent in its representation as
\acdefp{SDE}, enabling the utilization of powerful diffusion models \citep{song2019generative,
ho2020denoising}. Thus, the primary objective of this section is to establish that the minimizer of
\eqref{eq:gamma-IPF} can indeed be expressed as an \ac{SDE} with a readily available drift formula.

\newcommand{\forward}{v}%
\newcommand{\backward}{w}%

\newcommand{\optcontrol}{u^\star}

Before delving into the derivation, we first recall the important fact that the \ac{IPF} iterates can be expressed in terms of the \emph{time-reversal} of \acp{SDE}, which can be solved in practice via score matching techniques \citep{de2021diffusion,chen2021likelihood}. We provide a proof for this result for completeness.

\begin{restatable}{thm}{IPFSDE}
\label{thm:IPF-SDE}
Suppose that $\P^{n}$ is an \ac{SDE} given by
\begin{equation}\label{eq:Pn-sde}
  \dd X_t^n = \forward_t^n(X_t^n)\,\dd t + \sigma\dd W_t,\ X_0 \sim \mu_0,
\end{equation}
and that the time-reversal of $\P^{n + \half}$ is given by
\begin{equation}\label{eq:reversal-p-n-half}
  \dd Y^{n+\half}_t = \backward_{T - t}^{n+\half}(Y^{n+\half}_t)\,\dd t + \sigma\dd W_t, \ Y^{n+\half}_0 \sim \mu_T.
\end{equation}Then the drift vector field $\backward^{n+\half}_{t}$ satisfies:
\begin{equation}\label{eq:first-schrodinger-system-formula}
-\forward_{t}^n(x)  + \sigma^2\grad \log p^{n}_{t}(x) = \backward_{t}^{n+\half}(x)
\end{equation}where $p^n_t$ is the density of $\P^n_t$.
\end{restatable}
It is well-known that the condition \eqref{eq:first-schrodinger-system-formula} expresses precisely
the fact that \smash{$\P^{n + \half}$} is given by the {time-reversal} of \eqref{eq:Pn-sde}
\citep{follmer1985entropy,haussmannpardoux1986,song2020score}. 

In what follows, define the likelihood ratio to be $\lr^n_t \defeq  \nicefrac{p^{n+\half}_t}{ p^n_t}$. We are
now ready to present the main result of this section.

\newcommand{\ssstyle}{\scriptscriptstyle}

\begin{restatable}{thm}{gammaIPFSDE}
\label{thm:gamma-IPF-SDE}
Let $\P^n$ be given by the scheme \eqref{eq:gamma-IPF}, and let $\forward_t^n(\cdot)$
be the (forward) vector field corresponding to the \ac{SDE} representation of $\P^n$ in
\eqref{eq:Pn-sde}. Then $\forward_t^n(\cdot)$ satisfies the following recursive formula:
\begin{align}
\label{eq:gamma-IPF-SDE}
\tag{SDE$_\gamma$}
  \forward^{n+1}_t 
  = \forward^n_t + \gamma\sigma^2 \grad \log \lr^n_t
    - \sigma^2\nabla V_t,%
\end{align}where 
\begin{equation} \label{eq:value}
  V_t(x) = -\log \ex^{t,x}\brk*{\exp\prn*{-\frac{\sigma^2\gamma(1-\gamma)}{2} \int_t^T
\nrm*{\grad \log \lr^n_s(Y_s)}^2\,\dd s}},
\end{equation}
and the expectation is with respect to the law of the \ac{SDE} $(Y_s)_{s\geq t}$ starting at 
$Y_t = x$ and following
\begin{equation}\label{eq:uncontrolled-sde}
\dd Y_s = \crl*{\forward^n_s(Y_s) + \gamma \sigma^2 \grad \log \lr^n_s(Y_s)}\,\dd s + \sigma\dd W_s.
\end{equation}
\end{restatable}

When $\gamma = 1$, the $\nabla V_t$ term in \eqref{eq:gamma-IPF-SDE} disappears so that \cref{thm:gamma-IPF-SDE} is \cref{thm:IPF-SDE} applied twice, and we recover the iterative formula for the \ac{SDE} representation of \ac{IPF} \citep[\textbf{Proposition 4.2}]{pariset2023unbalanced}.

\begin{remark*}
Comparing \eqref{eq:gamma-IPF-SDE} with the classical \ac{IPF} iterates, we see that $\gamma$-\ac{IPF} \emph{can be efficiently implemented} provided we can calculate the additional $\nabla V_t$ component. In \cref{prop:ipf-as-soc}, we make this computational step possible by a establishing novel link to stochastic optimal control. As our paper primarily focuses on the theoretical understanding of the Sinkhorn and \ac{IPF} iterations, we defer the details to \cref{app:diffusion}.
\end{remark*}

\subsection{The Flow of Schr\"odinger \acp{SDE}}

In this section, we show how the results in \cref{sec:sinkhornflows} naturally lead
to \emph{a flow of \acp{SDE}}, \ie an evolution of path measures $(\P^s)_{s\geq 0}$ where each $\P^s$ is the law of an \ac{SDE} with certain drift $\forward^s_t(\cdot)$:
\begin{equation}
  \dd X^s_t = \forward^s_t(X^s_t)\dd t + \sigma\dd W_t.
\end{equation}

To streamline the exposition, we make the simplifying assumption that $T=\sigma=1$ and the reference measure $\refsde$ is given by the
law of the {reversible} Brownian motion $(\revbm_t)_{t\in [0, 1]}$ \citep{leonard2014}. Our conclusions remain applicable in the general case, but the notation becomes more cumbersome in that scenario.

Consider the static \ac{SB} problem in \eqref{eq:SB_BM}, which is nothing but \eqref{eq:eot-alt}
with cost function $c(x,y) = \frac{1}{2}\nrm{x-y}^2$ and $\eps = 1$. Recall its associated \schroeps
flow \eqref{eq:sink-schrodinger-pot-flow} defined via the Schr\"odinger potentials $f^s,g^s$.
Consider the path measures $\P^s$ defined by
\begin{equation} \label{eq:fg-transform}
  \rnder{\P^s}{\refsde} = \exp\crl{(f^s \oplus g^s)(\revbm_0, \revbm_1)}.
\end{equation}
Similar to the static case, these path measures are known to solve the \ac{SB} problem for their
corresponding marginals $\mu_0^s, \mu_1^s$ \citep{leonard2014} and, by construction, $\mu_0^s =
\mu_0$ for all $s$. 

We can now formally define an evolution of the path measures $\P^s$, where at each time $s$, $\P^s$
admits an \ac{SDE} representation which can be described using the Schr\"odinger potentials
$f^s,g^s$: For each $s$, define the function on $[0,1]\times\mathbb{R}^d$ by
\begin{align}
\label{eq:gt-def}
  g_t^s(z) &\coloneqq \log \ex \brk*{ e^{g^s(\revbm_1)} \mid \revbm_t = z}
\end{align}
so that $g_1^s \equiv g^s$. Then \citet[\textbf{Prop.~6}]{leonard2014} implies that $\P^s$ is the
law of the SDE:
\begin{equation}
\label{eq:sch-flow-sde}
  \dd X_t^s = \nabla g_t^s(X_t^s)\dd t + \dd W_t, \quad X_0^s \sim \mu_0.
\end{equation}
As a result, the mapping $s\mapsto (g_t^s)_{t\in[0,1]}$ %
can be regarded as the \emph{dynamic} \schroeps flow associated with $(g^s)_{s\geq 0}$, while
\eqref{eq:sch-flow-sde} can be considered as the continuous-time limit of the \ac{SDE}
representation of \eqref{eq:gamma-IPF}, as $\gamma \to 0$.

\section{Conclusions and future work}
\label{sec:conclusion}

In summary, our work introduces the continuous-time Sinkhorn algorithm as a novel approach to design
schemes that maintain convergence in the presence of noise and bias. It also unifies previously
isolated dynamics through the mirror descent perspective. We extend these insights to Schr\"odinger
bridges and the IPF procedure. These findings open doors to exciting future research directions,
including exploring connections with existing dynamics and the potential for achieving acceleration through momentum terms. %

\bibliographystyle{plainnat}
\bibliography{mrk,IEEEabrv,bibtex/Bibliography-PM,bibtex/Bibliography-YPH}

\appendix

\numberwithin{equation}{section}		%
\numberwithin{lemma}{section}		%
\numberwithin{proposition}{section}		%
\numberwithin{theorem}{section}		%
\numberwithin{corollary}{section}		%
\numberwithin{assumption}{section}	
\numberwithin{thm}{section}		%

\section{Background}
\label{app:background}

\subsection{Notions of Derivative}
Throughout this paper, we use the notation $\delta$ to denote the \emph{First Variation} operator.
In this section, we bring the necessary background for defining and using this operator. Our
discussion mostly follows \citep{aubin2022mirror} with a slight change of notation.

Let $\meas$ be a vector space of (signed) finite measures on the space $\Xsp$.
\newcommand{\derG}{\grad_{\text{G\^at}}}
\newcommand{\derF}{\grad_{\text{Fr\'e}}}
\begin{definition}[G\^ateaux and Fr\'echet Differentiability] 
  A functional $F$ is called G\^ateaux differentiable at $\mu \in \meas$, if 
  there exists a linear operator $\derG F(\mu)$ such that for any direction $\nu \in \meas$, one has
  \[
    \derG F(\mu)[\nu] = \lim_{h \to 0} \frac{F(\mu + h\nu) - F(\mu)}{h}.
  \]
  If the limit above holds uniformly in the unit ball in $\meas$, the function $F$ is called
  Fr\'echet differentiable, and we denote the resulting Fr\'echet derivative as $\derF F(\mu)$.
\end{definition}
The issue with the aforementioned definitions is that the limit must exist in all directions, which
means that the points of differentiability must be within the functional's domain, as stated.
Nevertheless, in the case of functionals like the relative entropy, the domain of $F$ has an
empty interior, as discussed by \citet{aubin2022mirror}.

For this, \citet{aubin2022mirror} propose to use the weaker notion of \emph{directional
derivative}, defined as follows:
\begin{definition}[Directional Derivative]
  For a functional $F$ and $\mu \in \meas$ define the directional derivative of $F$ at $\mu$ in the
  direction of $\nu \in \meas$ as 
  \[
    \ddv{F}{\mu}{\nu} = \lim_{h \downarrow 0} \frac{F(\mu + h\nu) - F(\mu)}{h}.
  \]
\end{definition}
See \citep[Remark 1]{aubin2022mirror} for a discussion on when this notion of derivative exists.
Specifically, for convex and proper functions (such as relative entropy) this notion of derivative
exists.

We are now ready to recall the notion of first variation, see \citep[Definition 2]{aubin2022mirror}
and the discussion afterward for a more in-depth exposition.
\begin{definition}[First Variation]
  Let $\cst$ be a subset of $\meas$. For a functional $F$ and $\mu \in \cst \cap \dom(F)$ define the
  first variation of $F$ at $\mu$ to be the element $\delta_\cst F(\mu) \in \meas^*$, where
  $\meas^*$ is the topological dual of $\meas$, such that it holds for all $\nu \in \cst \cap
  \dom(F)$ and $\xi = \nu - \mu \in \meas$:
  \[
    \inner{\delta_\cst F(\mu), \xi} = \ddv{F}{\mu}{\xi}.
  \]
  Here, $\inner{\cdot,\cdot}$ is the duality pairing of $\meas$ and $\meas^*$.
\end{definition}

In our exposition, we exclusively use the notation $\delta$ to denote the first variation
$\delta_\cst$ above for the respective constraint set.

\subsection{Selection of Results regarding \acp{SDE}}
The first important result is the time-reversal formula for diffusions. The result, along with the
necessary conditions can be found in \citep[\eg][]{haussmannpardoux1986}.
\begin{theorem}[Time-Reversal of Diffusions]\label{thm:time-reversal}
  Let $(X_t)_{t\in[0,1]}$ be the (strong) solution of $dX_t= \forward_t(X_t)\,dt + \sigma dW_t$, and
  assume $X_t$ has density $p_t$. Define the vector field
  \[
    \backward_t(x) = -\forward_{1-t}(x) + \sigma^2\grad \log p_{1-t}(x)
  \]
  and the operator $L_t$ which, when evaluated on $f \in C^\infty_c(\R^d)$, gives
  \[
    (L_t f)(x) = \frac{\sigma^2}{2}\lap f(x) + \inner{\backward_t(x), \grad f(x)}.
  \]
  Then, under some assumptions (see, \eg (A) in \citep{haussmannpardoux1986}), the process
  $(\hat{X}_t \coloneqq X_{1-t})_{0\leq t < 1}$ is a Markov diffusion process with generator $L_t$.
\end{theorem}

The following two theorems are special cases of a general Girsanov formula 
\citep[see \eg][Theorem 2.3]{leonard2011}.
\begin{theorem}[Girsanov]\label{thm:girsanov}
  Let $\P$ be the law of the semi-martingale
  \[
    X_t = X_0 + \int_0^t b_s\,ds + \sigma W_t,
  \]
  and suppose we are given a probability measure $\Q$ with $\KL{\Q \from \P} < \infty$. Then, there
  exists an $\R^d$-valued adapted process $\beta_t$ with  $\ex_{\Q}\brk*{\int_0^1
  \nrm{\beta_t}^2\,dt} < \infty$, such that $X$ has the semi-martingale decomposition
  \[
    X_t = X_0 + \int_0^t (b_s + \beta_s) \,ds + \sigma W^{\Q}_t,\quad \Q\text{-a.s.,}
  \]
  where $W^{\Q}$ is a $\Q$-Brownian motion. Moreover,
  \[
    \frac{d\Q}{d\P} = \frac{d\Q_0}{d\P_0}(X_0)\cdot \exp\crl*{
      \int_0^1 \frac{1}{\sigma}\inner{\beta_t, dW_t}
         - \frac{1}{2} \int_0^1 \frac{\nrm{\beta_t}^2}{\sigma^2}\,dt
    }.
  \]
\end{theorem}

\begin{corollary}[Relative Entropy of Diffusions]\label{cor:rel-ent-diffusions}
  Let $\P$ and $\Q$ be two path measures, with $\KL{\Q \from \P} < \infty$. Moreover, assume that
  under $\P$, the canonical process $X$ has the semi-martingale decomposition
  \[
    X_t = X_0 + \int_0^t b_s\,ds + W^\P_t,
  \]
  where $W^\P$ is a $\P$-Brownian motion, and under $\Q$,
  \[
    X_t = X_0 + \int_0^t c_s\,ds + W^\Q_t,
  \]
  where $W^\Q$ is a $\Q$-Brownian motion. Then, 
  \[
    \KL{\Q \from \P} = \ex_\Q\brk*{\log \frac{d\Q}{d\P}} = \KL{\Q_0 \from \P_0} +
    \ex_\Q\brk*{\frac{1}{2} \int_0^T \nrm{c_t - b_t}^2\,dt}.
  \]
\end{corollary}

\section{Proofs of \cref{sec:sinkhornflows}}
\label{app:proofs-sinkhorn}

\subsection{Results about the \stepsinkeps Iterates}
Recall from \cref{def:sinkhorn-step-md} that the \stepsinkeps iterates are defined via the recursion
\[
  \pi^{n+1}_{\gamma} = \argmin_{\pi \in \cst} 
  \crl*{ \inner{\delta\obj(\pi^{n}_{\gamma}),\pi - \pi^{n}_{\gamma}} + 
      \frac{\breg{\pi \from \pi^{n}_{\gamma}}}{\gamma_n}},\qquad \pi^0_\gamma = \refmeas.
\]
Using the values of $\delta \obj$ and $\breg{\cdot \from \cdot}$ from
\citep{aubin2022mirror}, this is equivalent to 
\begin{equation}\label{eq:sinkhorn-step-md-spelled-out}
  \pi^{n+1}_{\gamma} = \argmin_{\pi \in \cst} 
  \crl*{ \int \dd(\pi - \pi^n_\gamma)\log\rnder{\margy{(\pi^{n}_{\gamma})}}{\nu} +
      \frac{\KL{\pi \from \pi^{n}_{\gamma}}}{\gamma_n}},\qquad \pi^0_\gamma = \refmeas.
\end{equation}
  
\closedformstepsinkhorn*
\begin{proof}
  For brevity, we drop the $\gamma$ in $\pi^n_\gamma$.
  First, observe that by the chain rule of the relative entropy,
  \[
    \KL{\pi \from \pi^n} = \KL{\margy\pi \from \margy\pi^n} + \int
    \dd\margy\pi^n(y)\,\KL{\pi^n(\cdot\mid y) \from \pi(\cdot \mid y)}
  \]
  As $\pi^{n+ \half}$ is the minimizer of the KL above among all couplings with $\Ysp$-marginal
  $\nu$, this means that for $\dd\margy\pi^n(y)$-almost sure $y$, we have $\pi^n(\cdot\mid y) 
  = \pi(\cdot \mid y)$. This implies that
  \begin{equation}\label{eq:rnder-pnhalf-to-pn-sink}
    \rnder{\pi^{n+\half}}{\pi^n} = \rnder{\margy\pi^{n+\half}}{\margy\pi^n} = \rnder{\nu}{\margy\pi^n}.
  \end{equation}

  From the computation above and \eqref{eq:sinkhorn-step-md-spelled-out}, we have 
  \begin{align*}
    F(\pi^n) + \inner{\delta\obj(\pi^{n}),\pi - \pi^{n}} + 
      \frac{\breg{\pi \from \pi^{n}}}{\gamma_n} 
    &= \int \dd \pi^n\log\rnder{\margy\pi^{n}}{\nu} + \int \dd(\pi - \pi^n)\log\rnder{\margy\pi^{n}}{\nu} +
      \frac{\KL{\pi \from \pi^{n}}}{\gamma_n} \\
    &= \int \dd\pi\log\rnder{\margy\pi^{n}}{\nu} +
      \frac{\KL{\pi \from \pi^{n}}}{\gamma_n} \\
    &= \int \dd\pi\log\rnder{\pi^{n}}{\pi^{n+\half}} +
    \frac{\KL{\pi \from \pi^{n}}}{\gamma_n} && \text{by \eqref{eq:rnder-pnhalf-to-pn-sink}} \\
    &= \int \dd\pi\log\crl*{\rnder{\pi^{n}}{\pi^{n+\half}} \cdot
    \prn*{\rnder{\pi}{\pi^n}}^{1/\gamma_n}} \\
    &= \frac{1}{\gamma_n}\int \dd\pi\log\crl*{\prn*{\rnder{\pi}{\pi^{n+\half}}}^{\gamma_n} \cdot
    \prn*{\rnder{\pi}{\pi^n}}^{1-\gamma_n}} \\
    &= \frac{1}{\gamma_n} \prn*{ \gamma_n \KL{\pi \from \pi^{n+\half}} + (1-\gamma_n)\KL{\pi \from
    \pi^n}}.\qedhere
  \end{align*}
\end{proof}

As a corollary to \cref{lem:closed-form-step-sinkhorn}, we have
\begin{corollary}[Closed-form of the \stepsinkeps Step]
  \label{cor:closed-form-step-sinkhorn-density}
  The \stepsinkeps iterates have the density
  \[
    \pi^{n+1}_\gamma(\dd x, \dd y) \propto \mu(\dd x) \, 
    \prn*{\pi^{n+\half}(\dd y \mid x)}^{\gamma_n} \prn*{\pi^n(\dd y \mid x)}^{1- \gamma_n}.
  \]
\end{corollary}
\begin{proof}
  Let us drop the $\gamma$ for brevity. From \cref{lem:closed-form-step-sinkhorn} we know that
  \[
    \pi^{n+1} = \argmin \crl*{\gamma_n \KL{\pi \from \pi^{n+\half}} + (1-\gamma_n) \KL{\pi \from
    \pi^n} : \margx\pi = \mu}
  \]
  By the chain rule of relative entropy, and considering couplings $\pi$ with $\margx\pi= \mu$, we
  have
  \begin{align*}
    &\gamma_n \KL{\pi \from \pi^{n+\half}} + (1-\gamma_n) \KL{\pi \from \pi^n} \\
    &= \text{constant} + \int \mu(\dd x)\, \int \pi(\dd y \mid x) \log \prn*{
      \prn*{\frac{\pi(y \mid x)}{\pi^{n+\half}(y \mid x)}}^{\gamma_n}
      \prn*{\frac{\pi(y \mid x)}{\pi^{n}(y \mid x)}}^{1-\gamma_n}
    }.
  \end{align*}
  Thus, for each $x$, we have to set $\pi(\cdot \mid x)$ to the minimizer of 
  \[ 
    H(\pi(\cdot \mid x)) - \int \pi(\dd y\mid x)\,\log\crl*{\pi^{n+\half}(y \mid x)^{1-\gamma_n}
    \pi^{n}(y \mid x)^{1-\gamma_n}},
  \]
  which by standard optimality conditions in calculus of variations, we see that
  \[
    \pi^{n+1}(\cdot \mid x) \propto \pi^{n+\half}(y \mid x)^{1-\gamma_n} \pi^{n}(y \mid
    x)^{1-\gamma_n},
  \]
  and the claim of the corollary follows.
\end{proof}

\formulaKantorovicStepSinkhorn*
\begin{proof}
For easier readability, we drop the $\gamma$ subscript of $f^n_\gamma$ and $g^n_\gamma$. We prove
this lemma by induction. For $n = 0$, \eqref{eq:gamma-sink-potentials} holds because of the
initialization of the iterations. Now suppose that 
$\frac{d\pi^n}{d\refmeas} = \exp\prn*{ f^n \oplus g^n }$. Then, 
\[
  \pi^{n+\half}(x, y) = \nu(y)\,\pi^n(x \mid y) 
  = \frac{\mu(x)\exp\prn*{f^n(x) - \frac{c(x,y)}{\eps}}\nu(y)}%
  { \int \mu(x')\exp\prn*{f^n(x') - \frac{c(x',y)}{\eps}}\,dx'}
  = \refmeas(x, y) \exp\prn*{f_{n+\half} \oplus g_{n+\half}},
\]
where $f^{n+\half} = f^n$, and 
\[
  g^{n+\half}(y) = -\log \int \mu(x)\exp\prn*{f^n(x) - \frac{c(x,y)}{\eps}}\,dx
    = - \log \frac{\pi^n_\Ysp(y)}{\nu(y)\exp(g^n(y))}
    = - \log \frac{\pi^n_\Ysp(y)}{\nu(y)} + g^n(y)
\]

Now compute
\[
  \pi^n(y \mid x) = \frac{\exp\prn*{g^n(y) - \frac{c(x,y)}{\eps}}\nu(y)}%
  { \int \nu(y')\exp\prn*{g^n(y') - \frac{c(x,y')}{\eps}}\,dy'}
  \eqqcolon \frac{1}{A^n(x)} \exp\prn*{g^n(y) - \frac{c(x,y)}{\eps}}\nu(y)
\]
and
\[
  \pi^{n+\half}(y \mid x) = \frac{\exp\prn*{g^{n+\half}(y) - \frac{c(x,y)}{\eps}}\nu(y)}%
  { \int \nu(y')\exp\prn*{g^{n+\half}(y') - \frac{c(x,y')}{\eps}}\,dy'}
  \eqqcolon \frac{1}{A^{n+\half}(x)} \exp\prn*{g^{n + \half}(y) - \frac{c(x,y)}{\eps}}\nu(y).
\]
Thus,
\[
  \pi^{n+\half}(y \mid x)^{\gamma}\pi^n(y \mid x)^{1-\gamma}
  = \frac{1}{A^n(x)^{1-\gamma}A^{n+\half}(x)^\gamma}\exp\prn*{\gamma g_{n+\half}(y) +
  (1-\gamma)g_n(y) - \frac{c(x,y)}{\eps}}\nu(y).
\]
Recall that 
$\pi^{n+1}(dx, dy) = \mu(dx) \frac{1}{Z^n(x)}\pi^{n+\half}(y \mid x)^{\gamma}\pi^n(y \mid
x)^{1-\gamma}$.
Putting the values computed above, and gathering all terms that only depend on $x$ into $f^{n+1}$ shows that 
\[
  \pi^{n+1} = \exp\prn*{ f^{n+1}(x) + g^{n+1}(y) - \frac{c(x,y)}{\eps}}\mu(x)\nu(y),
\]
where
\[
  g^{n+1} = \gamma g^{n+\half} + (1-\gamma) g^{n} = -\gamma\log \frac{\pi^n_\Ysp}{\nu}
  + \gamma g^n + (1-\gamma)g^n = g^n - \gamma \log \frac{\pi^n_\Ysp}{\nu}. \qedhere
\]
\end{proof}

\subsection{Results about the \sinkeps and \schroeps Flows}
\timeDerivativeSinkhorn*
\newcommand{\ddga}{\frac{\dd}{\dd\gamma}}
\begin{proof}
Define $\pi^{\half}(x,y) = \nu(y)\,\pi(x \mid y)$ and for $\gamma \geq 0$, by
\cref{cor:closed-form-step-sinkhorn-density}
\[
  \pi^\gamma(x,y) = \mu(x)\frac{1}{Z^\gamma(x)} \pi^{\half}(y\mid x)^\gamma \pi(y \mid x)^{1-\gamma},
\]
where $Z^\gamma(x) = \int \pi^{\half}(y\mid x)^\gamma \pi(y \mid x)^{1-\gamma} \,dy$ is the normalization
factor. Note that $\pi^\gamma(y\mid x) = \pi^\gamma(x,y) / \mu(x)$.

Our goal is to characterize the derivative $\ddga\big\lvert_{\gamma = 0} \log \pi^\gamma(y
\mid x)$ for all fixed $x$. For that, we compute 
\[
  \frac{1}{\gamma}(\log \pi^\gamma(y\mid x) - \log \pi(y\mid x)) = -\frac{1}{\gamma} \log
  Z^\gamma(x) + \log \frac{\pi^{\half}(y \mid x)}{\pi(y \mid x)}.
\]
Thus, we only have to compute the limit
\[
  \lim_{\gamma \downarrow 0} \frac{1}{\gamma}\log Z^\gamma(x) = \ddga\Big\lvert_{\gamma =
    0} \log Z^\gamma(x) = \frac{1}{Z^0(x)} \int \pi(y \mid x) \log \frac{\pi^{\half}(y \mid x)}{\pi(y
    \mid x)}\,dy = -\KL{\pi(\cdot \mid x) \from \pi^{\half}(\cdot \mid x)}.
\]
Thus, 
\[
  \ddga\Big\lvert_{\gamma=0} \log \pi^\gamma(y \mid x) = -\log \frac{\pi(y \mid x)}{\pi^{\half}(y \mid x)}
  + \KL{\pi(\cdot \mid x) \from \pi^{\half}(\cdot \mid x)}.
\]
Notice that since the first marginal of $\pi$ is assumed to be fixed (to $\mu$), we have
\[
  \ddga\Big\lvert_{\gamma=0} \log \pi^\gamma(y \mid x) = \ddga\Big\lvert_{\gamma=0} \log \pi^\gamma(x, y) = 
    -\log \frac{\pi(y \mid x)}{\pi^{\half}(y \mid x)}
      + \KL{\pi(\cdot \mid x) \from \pi^{\half}(\cdot \mid x)}
\]

We have the identity:
\[
  \log \frac{\pi(y \mid x)}{\pi^{\half} (y \mid x)} 
      = \log \frac{\margy\pi(y)}{\nu(y)} + \log \frac{\margx\pi^{\half}(x)}{\margx\pi(x)}.
\]
With this, we can rewrite the evolution above as
\begin{align*}
  \ddga\Big\lvert_{\gamma=0} \log \pi^\gamma(x, y) 
    &= -\log \frac{\margy\pi(y)}{\nu(y)} - \log \frac{\margx\pi^{\half}(x)}{\margx\pi(x)}
    + \int \pi(z \mid x)\prn*{\log \frac{\margy\pi(z)}{\nu(z)} + \log
    \frac{\margx\pi^{\half}(x)}{\margx\pi(x)}}\,dz \\
    &= -\log \frac{\margy\pi(y)}{\nu(y)} 
    + \int \pi(z \mid x)\log \frac{\margy\pi(z)}{\nu(z)}\,dz.
\end{align*}
This proves the first part of the proposition.

For the evolution of $g^\gamma$, recall the recursion in
\cref{lem:formula-kantorovic-step-sinkhorn}, and see that when $\gamma \to 0$, the limit of
$g^\gamma$ satisfies the evolution in the proposition. For the evolution of $f^\gamma$, observe that
\[
  \log \pi^\gamma = (f^\gamma \oplus g^\gamma) + \log \refmeas
\]
Thus,
\[
  \ddga \log \pi^\gamma(x, y) = \ddga f^\gamma(x) + \ddga g^\gamma.
\]
Replacing the formula for $\ddga \log \pi^\gamma(x, y)$ and $\ddga g^\gamma$ above, gives the
formula for $\ddga f^\gamma$.
\end{proof}

\formulafenchelconjugaterelentconstrained*
\begin{proof}
For $h \in \dualsp$, we know that
\[
  \bregpot^*(h) = \sup_{\pi \in \cst} \inner{\pi, h} - \bregpot(\pi).
\]
For brevity, define $k(x,y) \coloneqq -h(x,y) - \log \refmeas(x, y)$. Then, 
\[
  \bregpot^*(h) = -\inf_{\pi \in \cst} \inner{\pi, k} + H(\pi),
\]
where $H(\pi)$ is the entropy of $\pi$. Now defined the Lagrangian
\[
  L(\pi, \lambda, \psi) \coloneqq \inner{\pi, k} + H(\pi) + \lambda (\inner{\pi, 1} - 1) +
  (\inner{\pi, \psi} - \inner{\mu, \psi}), \qwith \lambda \in \R,\quad \psi: \R^d \to \R.
\]
For a fixed $\lambda$ and $\psi$, we have that the minimizer of the Lagrangian is
\[
  \pi(x,y) = e^{-k(x,y) - \psi(x) - \lambda - 1}.
\]
Moreover, we have that 
\begin{align*}
  \psi &= \argmax_\psi \iint dx\,dy\, e^{-k(x,y) - \psi(x) - \lambda - 1} \prn*{ k(x,y) - k(x,y) -
  \psi(x) - \lambda - 1 + \lambda + \psi(x) } - \int \psi(x)\mu(x)\,dx \\
  &= \argmax_\psi -\iint dx\,dy\, e^{-k(x,y) - \psi(x) - \lambda - 1} - \int \psi(x)\mu(x)\,dx \\
  &= \argmin_\psi \iint dx \prn*{\int dy\, e^{-k(x,y)}} e^{- \psi(x) - \lambda - 1} + \psi(x)\mu(x),
\end{align*}
which gives 
\[
  \psi(x) = -\log \mu(x) + \log \int e^{-k(x,y)}\,dy - \lambda - 1,
\]
implying that
\[
  \pi(x,y) = \frac{e^{-k(x,y)}}{\int e^{-k(x,y')}\,dy'}\, \mu(x)  = \frac{\refmeas(x,y)\, e^{h(x,y)}}{\int
  \refmeas(x,y')\, e^{h(x,y')}\,dy'} \, \mu(x). \qedhere
\]
\end{proof}

\begin{remark*}
An easy consequence of \cref{lem:formula-fenchel-conjugate-relent-constrained} is the following
generalization of the classical dual isometry of Bregman divergences present in \ac{MD} for
Euclidean spaces to our setting.
\begin{lemma}[Dual Isometry of Bregman Divergences]
\label{lem:duality-MD}
Let $\pi,\pi' \in \cst = \coupmu$ having densities with respect to $\pi^{\textup{ref}}_{\eps}$. Then we have
\begin{align} 
\KL{ \pi \from  \pi'} =  \breg{\pi \from \pi' }= \bregst{ \delta\bregpot(\pi') \from \delta\bregpot(\pi)}.
\end{align}
\end{lemma}
\begin{proof}%
Write $\pi = e^h \refmeas$ and $\pi' = e^{h'} \refmeas$. We first note that, by
\eqref{eq:extremizer} and the fact that $\delta \bregpot(\pi) = h$, we have
\begin{align}
\nn
\delta\bregpot^*( \delta\bregpot(\pi))(x,y) &= \frac{\refmeas(x,y)\, e^{h(x,y)}}{\int \refmeas(x,y')\, e^{h(x,y')}\,dy'} \, \mu(x) \\
\nn
&= \frac{ \pi(x,y) }{\int \pi(x,y') \,dy'} \, \mu(x) \\
&= \pi(x,y) \label{eq:recover}
\end{align}where the last equality follows from $\pi \in \cst$. Similarly, we have $\delta\bregpot^*( \delta\bregpot(\pi')) = \pi'$. Using the formula for $\bregpot^*$ in \cref{lem:formula-fenchel-conjugate-relent-constrained}, we have
\begin{align}
\nn
\bregpot^*( h ) &= \inner{ \pi , h} - \KL{\pi\from \refmeas} \\
\nn
&= \int   \pi \log \frac{e^{h}\refmeas}{\refmeas} - \KL{\pi\from \refmeas} \\
\nn
&= \int  \pi \log \frac{\pi}{\refmeas}- \KL{\pi\from \refmeas}  \\
&= 0, \label{eq:null-bregpotst}
\end{align}and the exact same computation shows that $\bregpot^*( h' )=0$. We therefore have
\begin{align}
\nn
\bregst{ \delta\bregpot(\pi') \from \delta\bregpot(\pi)} &= \bregst{ h' \from h} \\
\nn
&= \bregpot^*(h') - \bregpot^*(h) - \inner{ \delta\bregpot^* \prn*{h}, h' - h   }\\
&= \inner{\pi, h - h'   } \label{eq:dual-norm}\\
\nn
&= \KL{\pi\from\pi'}
\end{align}where the third equality follows from \eqref{eq:recover} and \eqref{eq:null-bregpotst}.
\end{proof}
\end{remark*}

\mflow*
\begin{proof}
It is clear that \eqref{eq:mirror-flow-alt-measures} is equivalent to
\eqref{eq:mirror-flow-measures}. Before stating the proof of the theorem, we state some useful facts
about the Bregman potential and the first variation of its convex dual.

By the formula for $\delta \bregpot^*$ in \cref{lem:formula-fenchel-conjugate-relent-constrained},
we see that for any $h\in\dualsp$ and any $f:\Xsp\to\R$, we have $\delta\bregpot^*(h + f) =
\delta\bregpot^*(h)$:
\begin{align*}
\delta\bregpot^*(h + f)(x,y) 
= \mu(x) \frac{\refmeas(x,y)e^{h(x,y) + f(x)}}{\int\refmeas(x,y')
  e^{h(x,y') + f(x)}\,\dd y'}
= \mu(x) \frac{e^{f(x)}\refmeas(x,y)e^{h(x,y)}}{e^{f(x)}\int\refmeas(x,y')
  e^{h(x,y')}\,\dd y'}
= \delta\bregpot^*(h)(x,y),
\end{align*}
that is, if two functions in dual space $\dualsp$ only differ by a function of $x$, they correspond
to the same primal point. 

Consider a coupling $\pi\in\Pi_{c,\eps} \cap \cst $ written as $\rnder{\pi}{\refmeas} = \exp(f\oplus
g)$. By the definition of the Bregman potential $\bregpot(\pi) = \KL{\pi \from \refmeas}$, we see
that 
\[
  \delta\bregpot(\pi) = \log \rnder{\pi}{\refmeas} = f \oplus g.
\]
From the discussion above, $\pi = \delta\bregpot^*(\delta \bregpot(\pi)) = \delta\bregpot^*(f \oplus
g) = \delta\bregpot^*(g)$.

Now, consider the flow \eqref{eq:mirror-flow-measures} of $\dualvar^t_\eps$, started at $h^0_\eps =
\delta\bregpot(\primalvar^0_\eps) \eqqcolon f^0_\eps \oplus g^0_\eps$. Note that since $\delta
\obj(\cdot)$ is only a function of $y$, we have
\[
  \dualvar^t_\eps = f^0_\eps \oplus g^t_\eps.
\]
Noticing that (by construction) $\primalvar^t_\eps = \delta\bregpot^*(\dualvar^t_\eps) \in \Pi_{c,\eps}\cap \cst$ for
all $t$, our previous discussion implies
\[
  \primalvar^t_\eps = \delta\bregpot^*(\dualvar^t_\eps) = \delta\bregpot^*(f^0_\eps + g^t_\eps) =
  \delta\bregpot^*(g^t_\eps).
\]
Thus, in the evolution \eqref{eq:mirror-flow-measures}, if one only looks at the $y$ variable, one
gets
\[
  \ddt g^t_\eps = -{\delta F}(\primalvar^t_\eps),
\]
which is exactly the \schroeps flow.
\end{proof}

\sinkepsrate*
\begin{proof}
  
For brevity, we drop the $\eps$. First, we show that the objective function $\obj$ is decreasing
along the flow:
\begin{align*}
  &\ddt \obj(\pi^t) \\
  &= \inner*{ \delta \obj(\pi^t), \pi^t \ddt \log \pi^t} \\
  &= -\iint \pi^t(x,y) \prn*{\log \rnder{\margy\pi^t}{\nu}(y)}^2\,\dd x\,\dd y
     + \iint \pi^t(x,y) \log \rnder{\margy\pi^t}{\nu}(y) 
            \int \pi^t(z \mid x) \log \rnder{\margy\pi^t}{\nu}(z)\,\dd z \, \dd x\, \dd y
\end{align*}
Defining $k(x) = \int \pi^t(y \mid x) \log \rnder{\margy\pi^t}{\nu}(y)\,\dd y$, we see that the
second term above writes
\begin{align*}
  \iint \pi^t(x,y) \log \rnder{\margy\pi^t}{\nu}(y) 
            \int \pi^t(z \mid x) \log \rnder{\margy\pi^t}{\nu}(z)\,\dd z \, \dd x\, \dd y
            &= \iint \pi^t(x,y) \log \rnder{\margy\pi^t}{\nu}(y) 
            k(x)\,\dd x\,\dd y \\
            &= \iint \mu(x)\pi^t(y\mid x) \log \rnder{\margy\pi^t}{\nu}(y) 
            k(x)\,\dd x\,\dd y \\
            &= \int \mu(x)k(x)^2\,\dd x
\end{align*}
Now, by Jensen inequality, we have
\begin{align*}
  \int \mu(x) k(x)^2\,\dd x &= \int \mu(x) \prn*{
    \int \pi^t(y \mid x) \log \rnder{\margy\pi^t}{\nu}(y)\,\dd y
  }^2 \,\dd x \\
  & \leq \int \mu(x) 
    \int \pi^t(y \mid x) \prn*{\log \rnder{\margy\pi^t}{\nu}(y)}^2\,\dd y
  \,\dd x \\
  & = \iint \pi^t(x, y) \prn*{\log \rnder{\margy\pi^t}{\nu}(y)}^2\,\dd y
  \,\dd x.
\end{align*}
We thus have shown that 
\[
  \ddt \obj(\pi^t) \leq 0.
\]

For a Schr\"odinger potential $g$ and its corresponding coupling
$\pi = \getbackcoup{\eps}[g]$, define 
\[
  L(g) = \bregst{g \from \opt g} 
  = \bregpot^*(g) - \bregpot^*(\opt g) - \inner{\delta \bregpot^*(\opt g), g - \opt g} 
  = \bregpot^*(g) - \bregpot^*(\opt g) - \inner{\opt \pi), g - \opt g} 
\]
and observe that $\delta L(g) = \delta \bregpot^*(g) - \delta \bregpot^*(\opt g) = \pi - \opt\pi$.
We treat $L$ as a Lyapunov function of the \schroeps flow. For that, we compute
\begin{align}
\label{eq:hold1}
  \ddt L(g^t) &= \inner{ \delta L(g^t), \ddt g^t }
  = -\inner{ \pi^t - \opt\pi, \delta \obj (\pi^t) }
  \leq \obj(\opt\pi) - \obj(\pi^t),
\end{align}
where the inequality is due to convexity of $\obj$. Thus,
\[
  L(\pi^t) - L(\pi^0) = \int_0^t \dds L(\pi^s)\,\dd s \leq \int_0^t \obj(\opt\pi) -
  \obj(\pi^s)\,\dd s \leq t(F(\opt\pi) - F(\pi^t)),
\]
where the last inequality is due to the fact that $F(\pi^t)$ is non-increasing. Using the fact that
$L \geq 0$, we obtain the result.
\end{proof}

\subsection{Guarantees on Noisy \stepsinkeps}
\gammaconverg*

\begin{proof}%

\newcommand{\subs}{\leftarrow}

Since $F$ is convex and 1-smooth relative to $\bregpot$ \citep{aubin2022mirror}, \citet[\textbf{Lemma 5.2}]{hanzely2021fastest} with $\mu = 0$, $L_t \subs \frac{1}{\gamma}$, and $x\subs \opt{\pi}_\eps$ gives
\begin{align}
\ex \brk*{  F(\pi^{n+1}_\gamma) | \mathcal{F}_n } 
&\leq \frac{1}{\gamma} \prn*{ \KL{\opt{\pi}_\eps\from \pi^{n}_\gamma} -  \ex\brk*{\KL{\opt{\pi}_\eps\from \pi^{n+1}_\gamma } | \mathcal{F}_n }} +\gamma\sigma^2
\end{align}where $ \filter_n$ denotes the filtration generated by the stochastic algorithm up to step $n$. Taking expectation on both sides and summing over $n$, we get
\begin{align}
\frac{1}{n}\sum_{k=0}^{n-1} \ex\brk*{ F(\pi^{k}_\gamma)} \leq \frac{ \KL{\opt{\pi}_\eps\from \refmeas}  }{\gamma n} + \gamma\sigma^2.
\end{align}
The proof follows by using the convexity of $F(\cdot) \defeq \KL{\cdot \from \nu}$. \qedhere

\end{proof}

\para{A formal statment for \cref{thm:sinkapt}}

We will now present a formal theorem addressing the convergence of the method described in \eqref{eq:sinkhorn-step-update}, taking into account a noisy and biased oracle denoted as $\delta\obj$. To accomplish this, we will rely on the framework of \emph{stochastic approximation} \citep{Ben99, karimi2022dynamical, mertikopoulos2023unified}. Although this approach is well-established, it involves some technical complexities, so we have chosen to defer the details to this appendix.

Let $(\pi^n)_{n\in\mathbb{N}}$ be the sequence of measures generated by \eqref{eq:sinkhorn-step-update} with a noisy oracle $\tilde{\delta}\obj$ and step-sizes $(\gamma_n)_{n\in\mathbb{N}}$, and let $(\curr)_{n\in\mathbb{N}}$ be its corresponding Schr\"odinger potentials. We first define the ``effective time'' $\curr[\efftime]$ to be $\curr[\efftime] \defeq \sum_{\runalt=\start}^{\run} \step_\runalt$, which is the time that has elapsed at the $\run$-th iteration of the discrete-time process $\curr[\state]$. Using $\curr[\efftime]$, we consider the \emph{continuous-time interpolation} $\apt{\ctime}$ of $\curr$:
\begin{equation}
\label{eq:interpolation}
\apt{\ctime}
	\defeq \curr
		+ \frac{\ctime - \curr[\efftime]}{\next[\efftime] - \curr[\efftime]} (\next - \curr).
\end{equation}
Note that each $g(t)$ is a function on $\Ysp$.%
~The following assumption is standard in the stochastic approximation literature:
\begin{assumption}
\label{asm:technical}
Let $\pi^n$ and $\curr$ be given as above. We assume that \begin{enumerate*}[label=(\roman*)]
\item $\delta \obj$ is Lipschitz and bounded on a neighborhood of $(\pi^n)_{n\in\mathbb{N}}$, and
\item $(g(t))_{t\geq 0}$ is a \textbf{precompact} set in the topology of $L^\infty$.
\end{enumerate*}
\end{assumption}

It is worth highlighting that \cref{asm:technical} is a relatively mild technical condition that finds applicability in a wide range of practical scenarios. For example, it remains satisfied when employing bounded and H\"{o}lder continuous neural networks to parameterize distributions with compact support.

We are now ready to state the formal version of \cref{thm:sinkapt}.

\begin{thm}\label{thm:sinkapt-full}
Let $\pi^n$ and $\curr$ be given as above such that \cref{asm:technical} holds. Suppose that the step-size rule $\gamma_n$ is such that $\sum\gamma_n
= \infty$ and $\sum\gamma^2_n < \infty$. Denote by $\filter_n$ the filtration of the stochastic
algorithm up to iteration $n$, and its {martingale noise} and {bias} by
\begin{align*}
\curr[\bias]\defeq \exof{  \tilde{\delta}\obj(\pi^n) \given \filter_n} - \delta\obj(\pi^n),\\ %
\curr[\noise] \defeq \tilde{\delta}\obj(\pi^n)  -  \exof{  \tilde{\delta}\obj(\pi^n) \given \filter_n}.
\end{align*}
Then $\mathbb{P}(\lim_{n\to \infty}\pi^n = \opt{\pi}) = 1$ if the following holds almost surely:
\begin{align}
\label{eq:assumptions}
\lim_{n\to \infty} \Nrm*{ \curr[\bias]}_\infty = 0, \qand
\sup_n \exof*{  \Nrm*{\curr[\noise] }^2_{\infty} } \leq \sigma^2 < \infty.%
\end{align}
\end{thm}
\begin{proof}

The proof of the theorem is established through a combination of well-established results in stochastic approximation theory with our continuous-time framework. 

\cref{asm:technical} and \eqref{eq:assumptions} ensure that $\apt{\cdot}$ is a precompact \acli{APT}
of the associated continuous-time \schroeps flow given in \eqref{eq:sink-schrodinger-pot-flow}; see
\eg \cite[\textbf{Proposition 4.1}]{Ben99}. It follows from this association that the iterates
$(\curr)_{\run\geq 0}$ converge almost surely to an \acdef{ICT} set of the \schroeps flow. On the
other hand, within the course of our proof for \cref{thm:sinkepsrate}, we have established the
existence of a Lyapunov function for the \schroeps flow; see \eqref{eq:hold1}. Consequently, the
only possible \ac{ICT} set is identified as the singleton set $\{\opt{g}_\eps\}$
\citep[\textbf{Proposition 6.4}]{Ben99}. This, in turn, implies that the following event happens
almost surely:
\begin{align*}
\lim_{n\to\infty} \pi^n &= \lim_{n\to\infty}\delta \obj^*(\curr)\\
&=\delta \obj^*(\opt{g}_\eps) \\
&= \opt{\pi}_\eps.\qedhere
\end{align*}

\end{proof}

\section{On Schr\"odinger Bridges}
\label{app:diffusion}

\subsection{Proof of \cref{prop:ipf-is-md}}
First we need to bring some lemmas.

\begin{lemma}\label{lem:formula-dir-der-kl}
  Suppose $\P \in \prob(\Omega)$ and $\Q \in \meas(\Omega)$ be a finite measure. For $\obj(\P) =
  \KL{\P_T \from \mu_T}$ we have
  \[
    \ddv{\obj}{\P}{\Q} = \int_{\R^d} \dd\Q_T \log \rnder{\P_T}{\mu_T}.
  \]
\end{lemma}
\begin{proof}
  As the function $\obj$ only depends on the marginals at time $1$, the claim follows from a similar
  calculation as in \citep[Prop. 5]{aubin2022mirror}.
\end{proof}

\begin{lemma}\label{lem:formula-breg-kl}
  For $\bregpot(\P) = \KL{\P \from \refsde}$, we have $\breg{\P \from \Q} = \KL{\P \from \Q}$.
\end{lemma}
\begin{proof}
  Similar to \citep[Example 2]{aubin2022mirror}.
\end{proof}

\begin{lemma}\label{lem:radon-nikodym-iters-ipf}
  For the iteration \eqref{eq:ipf}, it holds that
  \[
    \rnder{\P^{n+ \half}}{\P^n} = \rnder{\mu_T}{\P^n_T}.
  \]
\end{lemma}
\begin{proof}
  By the chain rule of KL divergence, we know that
  \[
    \KL{\P \from \P^n} = \KL{\P_T \from \P^n_T} 
      + \int \KL{ \P(\cdot \mid X_T = x) \from
            \P^n(\cdot \mid X_T = x)}\, \dd\P_T(x)
  \]
  Notice that in the first part of the iteration \eqref{eq:ipf}, the last marginal is fixed, thus,
  the first term above is fixed, and the minimizer shall be
  \[
    \P^{n+ \half}(\cdot) = \int \dd\mu_T(x)\, \P^n(\cdot \mid X_T = x).
  \]
  From this representation, the claim of the lemma is clear.
\end{proof}

We can now go ahead and prove \cref{prop:ipf-is-md}, stated below for convenience.
\ipfismd*
\begin{proof}
  For a path measure $\P$, compute the following:
  \begin{align*}
    &\obj(\P^n) + \ddv{\obj}{\P^{n}}{\P - \P^{n}} + \breg{\P \from \P^{n}} \\
    &= \KL{\P^n_T \from \mu_T} + \int_{\R^d} d(\P - \P^n)_T \log \frac{d\P^n_T}{d\mu_T} + \KL{\P
    \from \P^{n}}  && \text{by \cref{lem:formula-dir-der-kl,lem:formula-breg-kl}} \\
    &= \int_{\R^d} d\P_T \log \frac{d\P^n_T}{d\mu_T} + \KL{\P \from \P^{n}} \\
    &= \int_{\Omega} d\P \log \frac{d\P^n_T}{d\mu_T} + \KL{\P \from \P^{n}} \\
    &= \int_{\Omega} d\P \log \frac{d\P^n}{d\P^{n+\half}} + \KL{\P \from \P^{n}} 
        && \text{by \cref{lem:radon-nikodym-iters-ipf}} \\
    &= \int_{\Omega} d\P \log \crl*{\frac{d\P^n}{d\P^{n+\half}} \cdot \frac{d\P}{d\P^n}} \\
    &= \KL{\P \from \P^{n+\half}}.
  \end{align*}
  Now it is clear that the minimizer of the above in the set $\cst$ is exactly $\P^{n+1}$.
\end{proof}

\subsection{SDE Representation and the Drift Formula}
\IPFSDE*
\begin{proof}
First, letting $\hat{\P}^n$ to denote the law of the time-reversal of $\P^n$, observe that since the
time reversal of $\P^{n+\half}$ solves $\argmin \crl{  \KL{\P \from \hat{\P}^n} : \P_0 = \mu_T}$,
its SDE representation is the same as the one for $\hat{\P}^n$, and only its initial datum is set
to be $\mu_T$. By the time reversal formula (\cref{thm:time-reversal}), $\hat{\P}^n$ corresponds to 
\[
  dY^{n}_t = \crl*{-\forward_{T-t}^n(Y^{n}_t) + \sigma^2\grad \log p^{n}_{T-t}(Y^n_t)}\,dt + \sigma dW_t, \qquad Y^{n}_T \sim \mu_0,
\]
where $p^n_t$ is the density of $\P^n_t$. This means that this should coincide with the SDE for time
reversal of $\P^{n+\half}$, that is
\[
  -\forward_{t}^n(x)  + \sigma^2\grad \log p^{n}_{t}(x) = \backward_{t}^{n+\half}(x). \qedhere
\]
\end{proof}
\gammaIPFSDE*
\begin{proof}
Note that the path measure $\P^{n+\half}$ corresponds to the reversal of
\eqref{eq:reversal-p-n-half}, which is a process with drift $\forward^{n+\half} \coloneqq
-\backward^{n+\half}_t + \sigma^2\grad \log p^{n+\half}_t$, with $p^{n+\half}_t$ being the density
of $\P^{n+\half}_t$. Recall that $\P^{n+1}$ is the solution to the minimization 
\[
  \P^{n+1} = \argmin_{\P \in \cst} \crl*{ \gamma_n\, \KL{\P \from \P^{n+\half}}
  + (1 - \gamma_n)\,\KL{\P \from \P^n}},
\]
and suppose that it corresponds to the SDE
\begin{equation}\label{eq:basic-sde-of-minimizer-of-weighted-kl}
  \dd X^u_t = \prn*{b^\gamma_t(X^u_t) + u_t}\,\dd t + \sigma \dd W_t,
\end{equation}
with $X_0 \sim \mu_0$, where we define the drift $b^\gamma_t$ as
\[
  b^\gamma_t \coloneqq 
  \gamma \forward^{n+\half}_t + (1-\gamma)\forward^n_t = 
  \gamma \cdot (-\backward^{n+\half}_t + \sigma^2\grad \log p^{n+\half}_t)
  + (1-\gamma) \cdot \forward^n_t= \forward^n_t +\gamma\sigma^2 \grad \log \lr^n_t
\]
by \cref{thm:IPF-SDE}. The reason that we take \eqref{eq:basic-sde-of-minimizer-of-weighted-kl} as
an SDE representation of $\P^{n+1}$ is that, firstly, it should be a diffusion with the same
diffusion coefficient, and its drift shall be the ``weighted average'' of the drifts of
$\P^{n+\half}$ and $\P^n$, with some correction $u_t$.

It turns out that characterizing $u_t$ corresponds to solving a stochastic optimal control problem.
Concretely, by the Girsanov theorem (see \cref{cor:rel-ent-diffusions}), we obtain
\begin{align*}
  &\gamma \KL{\P \from \P^{n+\half}} + (1-\gamma)\KL{\P \from \P^{n}}  \\
  &\quad = \text{constant} +
  \ex_\P\brk*{\frac{1}{2\sigma^2}\int_0^T \crl*{\gamma \nrm{u_t + b^\gamma_t(X_t) -
      \forward^{n+\half}_t(X_t)}^2 +
  (1-\gamma) \nrm{u_t + b^\gamma_t(X_t) - \forward^n_t(X_t)}^2 }\,\dd t}\\
  &\quad = \text{constant} +
  \frac{1}{\sigma^2}\ex_\P\brk*{\frac{1}{2}\int_0^T \nrm{u_t}^2\,\dd t + \frac{\gamma(1-\gamma)}{2}\int_0^T
  \nrm{\forward^{n+\half}_t(X_t) - \forward^n_t(X_t)}^2\,\dd t} \\
  &\quad = \text{constant} +
  \frac{1}{\sigma^2}\ex_\P\brk*{\int_0^T \frac{1}{2}\nrm{u_t}^2\,\dd t + \frac{\sigma^4\gamma(1-\gamma)}{2}\int_0^T
  \nrm{\grad \log \lr^n_t(X_t)}^2\,\dd t}
\end{align*}
where the constant is the weighted sum of KL divergences for time marginals at 0, and is some
fixed number (as we fixed the initial distribution of $\P$). Now we recognize that the minimization
problem \eqref{eq:gamma-IPF} is a stochastic optimal control problem with running
cost
\[
  r(t, x,\alpha) = \frac{1}{2}\nrm{\alpha}^2 +
  \frac{\sigma^4\gamma(1-\gamma)}{2}\nrm{\grad\log\lr^n_t(x)}^2,
\]
the cost functional $J[u] = \ex_\P \brk*{ \int_0^t r(t, X^u_t, u_t)\,dt }$, and zero terminal
cost. By \cref{prop:stoch-opt-control-solution} below (setting $c_t = \frac{1}{2}\sigma^4\gamma(1-\gamma)\nrm{\grad\log\lr^n_t}^2$), the value function $V_t(x)$ of this control
problem is
\begin{align*}
  V_t(x) &= -\sigma^2\log \ex^{t,x}\brk*{\exp\prn*{-\frac{1}{\sigma^2}\int_t^T
           \frac{1}{2}\sigma^4\gamma(1-\gamma)\nrm{\grad\log\lr^n_s(Y_s)}^2\,\dd s}} \\
  &=-\sigma^2\log \ex^{t,x}\brk*{\exp\prn*{-\frac{\sigma^2\gamma(1-\gamma)}{2}\int_t^T
           \nrm{\grad\log\lr^n_s(Y_s)}^2\,\dd s}}
\end{align*}
where $\ex^{t,x}$ is expectation with respect to the law of the process $Y$, given that it starts at
$x$ at time $t$; $Y_t = x$, and the optimal control $u_t = -\grad V_t$.
Thus, denoting the drift of $\P^{n+1}$ as $\forward^{n+1}$, we see that $\P^{n+1}$ is the law of the SDE
\[
  \dd X_t = \forward^{n+1}_t(X_t)\,\dd t + \sigma \dd W_t, \quad X_0 \sim \mu_0,
\]
with
\begin{align*}
  \forward^{n+1}_t(x) 
  &= b^\gamma_t(x) + u_t(x) \\
  &= \forward^n_t(x) + \gamma \sigma^2 \grad \log \lr^n(x) + u_t(x) \\
  &= \forward^n_t(x) + \gamma \sigma^2 \grad \log \lr^n(x) 
    + \sigma^2 \grad \log \ex^{t,x}\brk*{
      \exp \prn*{-\frac{\sigma^2\gamma(1-\gamma)}{2} \int_t^T
        \nrm*{\grad \log \lr^n_s(Y_s)}^2\,\dd s }
    },
\end{align*}
where $(Y_s)_{s \geq t}$ follows the SDE
\[
  \dd Y_s = \crl*{\forward^n_s(Y_s) + \gamma \sigma^2 \grad \log \lr^n_s(Y_s)}\,\dd s + \sigma \dd W_s,
  \quad Y_t=x. \qedhere
\]
\end{proof}

We used the following result regarding computation of the value function of a specific stochastic
optimal control problem in \cref{thm:gamma-IPF-SDE}.
\begin{proposition}\label{prop:stoch-opt-control-solution}
  Let $b_t$ be a given drift, and consider the controlled SDE
  \[
    dX^u_t = (b_t(X^u_t) + u_t)\,dt + \sigma dW_t
  \]
  along with the following stochastic optimal control problem:
  \[
    \min_{u} J[u] \coloneqq \ex\brk*{ \int_0^T \tfrac{1}{2}\nrm{u_t}^2 + c_t(X^u_t)\,dt }.
  \]
  Then, the value function is given by
  \[
    V_s(x) = -\sigma^2 \log\ex^{s,x}\brk*{
      \exp \prn*{-\frac{1}{\sigma^2} \int_s^T c_t(Y_t)\,dt }
    },
  \]
  where $Y_t$ is the solution of the uncontrolled SDE $dY_t = b_t(Y_t)\,dt +\sigma dW_t$. Moreover,
  the optimal control is of feedback type $u_t(x) = -\grad V_t(x)$.
\end{proposition}
\begin{proof}
The value function $V_t(x)$ of this control problem shall satisfy the HJB equation, which writes
\[
  \partial_t V_t(x) + \min_{\alpha \in \R^d} \crl*{ 
  \inner{b_t(x) + \alpha, \grad V_t(x)} + \frac{\sigma^2}{2}\lap V_t(x) + \tfrac{1}{2}\nrm{\alpha}^2 +
  c_t(x)} = 0,\quad
  V_T(x) = 0,
\]
and evaluates to the optimal control $\alpha^*(t, x)= -\grad V_t(x)$. Replacing $\alpha$ with its
optimal value gives
\[
  \partial_t V_t(x) - \frac{1}{2}\nrm{\grad V_t(x)}^2 + \frac{\sigma^2}{2} \lap V_t(x) +
  \inner{b_t(x), \grad V_t(x)} + c_t(x) = 0, \quad V_T(x) = 0.
\]
Inspired by the Fleming log transform \citep{fleming1977}, make the change of variables $V_t(x) = -\sigma^2\log E_t(x)$, and observe that the equation above
becomes
\[
  \partial_t E_t(x) + \frac{\sigma^2}{2} \lap E_t(x) +
  \inner{b_t(x), \grad E_t(x)} = \frac{1}{\sigma^2}E_t(x)\,c_t(x) , \quad E_T(x) = 1.
\]
Following a similar argument as in \citep{prapavon1990}, consider the uncontrolled diffusion process
$Y_t$ with generator $\frac{\sigma^2}{2}\lap + b_t \cdot \grad$. By the Feynman-Kac formula,
\[
  E_s(x) = \ex^{s,x}\brk*{\exp\prn*{-\frac{1}{\sigma^2}\int_s^T c_t(Y_t)\,dt}},
\]
where $\ex^{s,x}$ means expectation with respect to the law of the process $Y$, given that it starts
at $x$ at time $s$.
\end{proof}

\subsection{On the Implementation of the $\gamma$-IPF iteration}

Although our paper focuses on the theoretical understanding of the Sinkhorn and \ac{IPF} iterates,
we briefly remark that the formula in \cref{thm:gamma-IPF-SDE} admits a practically efficient
implementation. To see this, notice that the $ \grad \log \lr^n_t $ term in \eqref{eq:gamma-IPF-SDE}
is the standard \emph{Stein score} ratio that can be estimated by various diffusion models and is
present in most practical training procedures of \ac{SB}. On the other hand, the computation of the
additional term $V_t$ in \eqref{eq:gamma-IPF-SDE} is facilitated by the following connection to
stochastic optimal control, whose proof is already present in the proof of \cref{thm:gamma-IPF-SDE}.

\begin{proposition}\label{prop:ipf-as-soc}
  The minimization  \eqref{eq:gamma-IPF} is equivalent to solving the following stochastic optimal
  control  problem: Consider the following controlled \ac{SDE} with drift
  $b_t = \forward^n_t + \gamma \grad \log \lr^n_t$ and control $u_t$:
  \begin{equation} \label{eq:controled-sde}
    \dd X^u_t = (b_t(X^u_t) + u_t)\,\dd t + \sigma\dd W_t,
  \end{equation}
  and the cost functional
  \begin{equation} \label{eq:cost}
    J[u] \coloneqq \ex\brk*{ \int_0^T \frac{1}{2}\nrm{u_t}^2 + c_t(X^u_t)\,dt }
  \end{equation}
  with $c_t = \frac{1}{2}\sigma^2\gamma(1-\gamma)\nrm{\grad \log \lr^n_t}^2$.
  Then, the value function of the stochastic optimal control problem
  $\min_u J[u]$ is $\sigma^2 V_t$, where $V_t$ is defined in
  \eqref{eq:value} and the optimal control is $\optcontrol = -\sigma^2\grad V_t$.
\end{proposition}

It turns out that the value function $V_t$ can also be evaluated as an expectation that involves
only the law of the standard Brownian motion:
\begin{lemma}\label{lem:value-formula-wiener}
  The function $V_t$ in \eqref{eq:value} also satisfies
  \[
    V_t(x) = -\log\ex\brk*{\exp\prn*{\frac{1}{\sigma} \int_t^T\inner{
         b_s(x + \sigma W_{s-t}),\dd W_{s-t}}
    -\frac{1}{2\sigma^2}\int_t^T \nrm{b_s(x+\sigma W_{s-t})}^2 + c_s(x + \sigma W_{s-t})\,\dd s}},
  \]
  where the expectation is with respect to the standard Brownian motion $(W_t)_{t\geq 0}$.
\end{lemma}
\begin{proof}
  Proof is a simple application of the Girsanov theorem (\cref{thm:girsanov}), applied to the \acp{SDE} \eqref{eq:controled-sde} and $\sigma W_t$, where $W_t$ is a standard Brownian motion.
\end{proof}

\cref{prop:ipf-as-soc} and \cref{lem:value-formula-wiener} furnishes at least two different ways of
computing $V_t$ in \eqref{eq:gamma-IPF-SDE}. First, given $\grad \log \lr^n_t$ which is
given by the usual score matching procedure in \ac{SB} training, one can compute the value function
using standard approximation techniques in control theory for integration with respect  to standard
Brownian motion \citep{zhang2021path}.
Alternatively, another common practice is to connect the value function via the
\emph{Feynman-Kac formula} to \acp{SDE} with \emph{killing}. Concretely, since the cost $c_t$ above is non-negative, one can simulate the uncontrolled \ac{SDE} \eqref{eq:uncontrolled-sde}, and kill it at a rate
$\frac{1}{\sigma^2}c_t = \frac{\gamma(1-\gamma)}{2}\nrm*{\grad \log \lr^n_t}^2$, that is,
\[
  \mathbb{P}\brk*{ Y_{t+h} \text{ is killed} \mid Y_t}
  = \frac{\gamma(1-\gamma)}{2}\nrm*{\grad \log \lr^n_t(Y_t)}^2 + o(h).
\]
These procedures are already employed in the \ac{SB} community in other contexts
\citep{liu2022deep,pariset2023unbalanced}.

\end{document}